\pdfoutput=1

\documentclass[11pt]{article}

\usepackage[final]{acl}

\usepackage{times}
\usepackage{latexsym}

\usepackage[T1]{fontenc}

\usepackage[utf8]{inputenc}

\usepackage{microtype}

\usepackage{inconsolata}

\usepackage{kotex}
\usepackage{subcaption}
\usepackage{booktabs}
\usepackage{multirow}
\usepackage{makecell}
\usepackage{graphics}
\usepackage{amssymb}
\usepackage{bbm}
\usepackage{amsmath}
\usepackage{lipsum}
\usepackage{graphicx}
\usepackage[english]{babel}
\usepackage{algorithm}
\usepackage[noend]{algpseudocode}
\usepackage{enumitem}
\usepackage{bm}
\usepackage{mathtools}
\usepackage{amsthm}
\newtheorem{definition}{Definition}
\newtheorem{theorem}{Theorem}
\newtheorem{corollary}{Corollary}[theorem]
\makeatletter
\algnewcommand{\LineComment}[1]{\Statex \hskip\ALG@thistlm \(\triangleright\) #1}
\makeatother

\usepackage{xcolor}
\usepackage{multicol}

\usepackage{etoolbox}
\usepackage{listings}

\definecolor{middlegray}{rgb}{0.5,0.5,0.5}
\title{Unleashing Multi-Hop Reasoning Potential in Large Language Models through Repetition of Misordered Context}

\author{{\bf Sangwon Yu$^{1}$} \hspace{6mm}  {\bf Ik-hwan Kim$^{1}$} \hspace{6mm}  {\bf Jongyoon Song$^{1}$} \\  {\bf Saehyung Lee$^{1}$} \hspace{6mm}  {\bf Junsung Park$^{1}$} \hspace{6mm}  {\bf Sungroh Yoon$^{1,2}$\Thanks{\hspace{0.1em} Corresponding author}} \\ \\
   $^{1}$Department of Electrical and Computer Engineering, Seoul National University \\
   $^{2}$AIIS, ASRI, INMC, ISRC, and IPAI, Seoul National University \\ \\
   {\tt \fontsize{10}{10}\selectfont \{dbtkddnjs96, 12kimih, coms1580, halo8218, jerryray, sryoon\}@snu.ac.kr}}

\begin{document}
\maketitle
\begin{abstract}
Multi-hop reasoning, which requires multi-step reasoning based on the supporting documents within a given context, remains challenging for large language models (LLMs). LLMs often struggle to filter out irrelevant documents within the context, and their performance is sensitive to the absolute position of supporting documents within that context. In this paper, we identify an additional challenge: LLMs' performance is also sensitive to the order, relative position, in which the supporting documents are presented. We refer to this as the \textbf{misordered context} problem. To address this issue, based on the theoretical approach, we propose a simple yet effective method called \textbf{co}ntext \textbf{re}petition (\textbf{CoRe}), which involves prompting the model by repeatedly presenting the context. This ensures that certain contiguous reasoning segments within supporting documents are presented in the optimal order, effectively guiding the model's reasoning in the appropriate direction. Applying CoRe, we improve the F1 score by up to 30\%p on multi-hop QA tasks and increase accuracy by up to 70\%p on a synthetic task. Additionally, CoRe helps mitigate the well-known ``lost-in-the-middle'' problem in LLMs and can be effectively combined with retrieval-based approaches utilizing Chain-of-Thought (CoT) reasoning.
\end{abstract}

\section{Introduction} \label{main_intro}
Large language models (LLMs) (\citealp{gpt3}; \citealp{ouyang2022training}; \citealp{touvron2023llama}) are capable of performing various complex natural language processing tasks due to their emergent abilities in in-context learning, instruction following, and step-by-step reasoning (\citealp{wei2022cot}; \citealp{kojima2022zeroshotcot}; \citealp{zhao2023llmsurvey}). However, multi-hop reasoning still remains a challenging task for LLMs. In this task, multiple external source documents related to the given query are provided as context, and the model searches for useful supporting documents within this context to find the answer. The model derives an answer through multiple steps of reasoning based on these supporting documents (\citealp{mavi2024multi}).

\begin{figure}[t]
{
\begin{center}
\centerline{\includegraphics[width=\columnwidth]{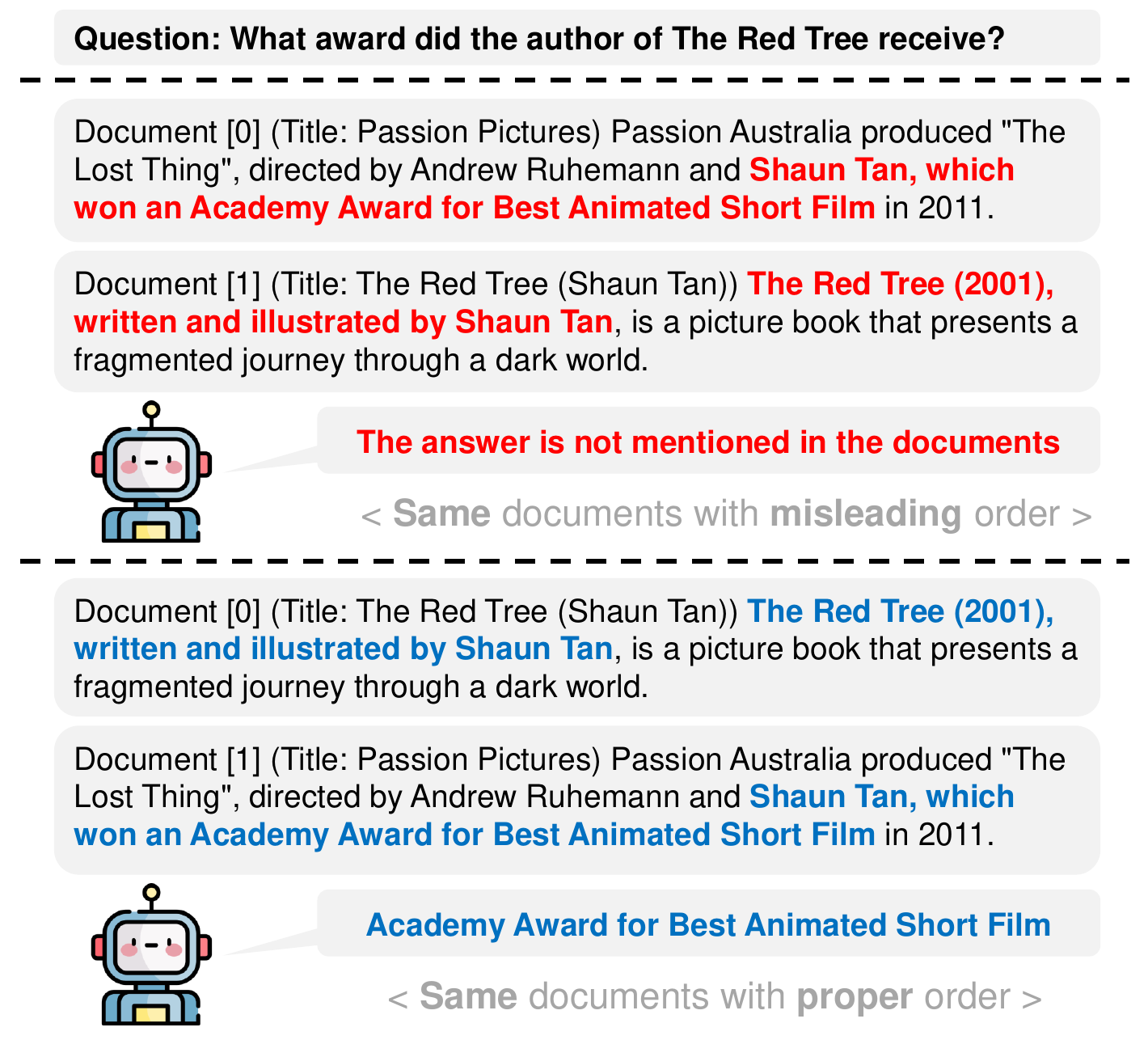}}
\caption{An example of the misordered context problem in multi-hop reasoning of large language models, sampled from the MuSiQue dataset. Model performance is sensitive to the order of given documents.}
\label{fig_intro}
\end{center}
}
\end{figure}

During this process, the model faces several challenges in terms of grounding to the context (\citealp{gao2023retrievalsurvey}). First, the model struggles to filter out noisy documents that are irrelevant to the correct answer (\citealp{shi2023large}). In practice, noisy documents can significantly degrade the reasoning performance of LLMs (\citealp{Cuconasu_2024_powerofnoise}; \citealp{wu2024how}). Additionally, LLMs are highly sensitive to the absolute position of supporting documents within the context. Specifically, when a document is positioned in the middle of the context, the model may fail to recognize it properly, resulting in a ``lost-in-the-middle'' problem (\citealp{liu2023lostmiddlelanguagemodels}). These issues become more critical as the maximum length of the context grows and the number of retrieved documents increases (\citealp{lee2024canlongcontext}; \citealp{li2024makinglongcoxtllmbettermultihop}).

In this study, we address additional context grounding problems related to multi-hop reasoning that have not been deeply discussed so far. In a multi-hop reasoning task, there are multiple supporting documents, and the model derives the answer by considering each document through step-by-step reasoning. Due to the nature of the decoder-only structure, which performs causal attention, LLMs can only perceive documents in a left-to-right order. If this order is presented unfavorably for the model’s reasoning, the model's performance may be significantly degraded. 
Following this rationale, we observe that, even in ideal situations without noisy documents, LLMs exhibit up to a 26\%p difference in F1 score depending on the order, relative position, of supporting documents. Since there is no guarantee that supporting documents will be presented to the model in the appropriate order for the model reasoning during actual tasks, it is inevitable that this problem arises. We call it the \textbf{misordered context} problem.  Figure \ref{fig_intro} shows an example of the misordered context problem.

To address the problem, we approach it from the perspective of context augmentation. We aim to construct the augmented context that presents the supporting documents in a more appropriate order than the original context. Through theoretical analysis, we demonstrate that by repeating the context containing $k$ supporting documents $k$ times, all possible orders of presenting the documents can be covered. That is, \textit{the $k$-repeated context can always present the supporting documents to the model in the optimal order}. In real-world scenarios, however, the value of $k$ for a given context is unknown. Furthermore, repeating the context leads to an increase in inference time costs. Based on this theoretical analysis and its identified limitations, we propose the \textbf{co}ntext \textbf{re}petition (\textbf{CoRe}) method: \textit{the given context is repeated $\hat{k}$ times in which $\hat{k}$ is the predetermined hyperparameter}. This practical approach ensures that all $\hat{k}$-contiguous reasoning chains within supporting documents are presented in the optimal order, guiding LLM's reasoning in a more coherent and structured manner.

To evaluate CoRe, we conduct experiments on 3 multi-hop QA benchmarks---HotpotQA (\citealp{yang2018hotpotqa}), 2WikiMultihopQA (\citealp{xanh2020_2wikimultihop}), and MuSiQue (\citealp{trivedi2021musique})---as well as an additional synthetic task. In the synthetic task, the model is required to search for elements in lists composed of non-negative integer elements, based on context where information is presented in a reverse order. In the evaluation, CoRe significantly improves multi-hop QA task performance across various LLMs. For example, in the 2WikiMultihopQA task using the Llama-3.1-8B-Instruct model, we observe up to 30\%p improvement in F1 score, and in the synthetic task, we observe up to a 70\%p improvement in accuracy. Through in-depth empirical analysis, we demonstrate that CoRe provides performance benefits to the model by improving order-related reasoning. Additionally, we show that CoRe alleviates the previously known ``lost-in-the-middle'' problem and can be effectively combined with Chain-of-Thought (CoT) methods in retrieval-augmented tasks.

In summary, our work (1) introduces the misordered context problem, a critical challenge for LLMs in multi-hop reasoning, (2) theoretically proposes context augmentation as a solution to this issue, which forms the foundation of our practical and effective CoRe method, and (3) demonstrates that CoRe significantly enhances model performance across various tasks and effectively mitigates the misordered context problem.

\section{Related Work} \label{main_related_work}
\begin{figure*}[!t]
    \centering
    \includegraphics[width=\textwidth]{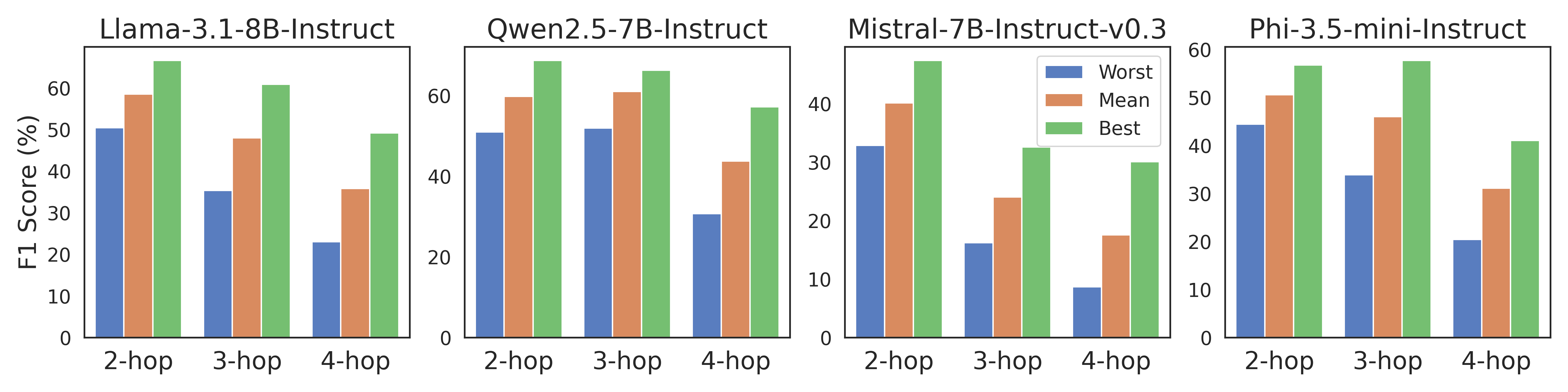} 
    \caption{
    Evaluation results of F1 score for each query type in MuSiQue with permuted clean contexts.
    }
    \label{fig_dc_problem}
\end{figure*}
The process of modeling the multi-step reasoning required in a multi-hop QA task typically involves methods that decompose the query step-by-step to predict the answer (\citealp{min2019multi}; \citealp{yadav2019quick}; \citealp{malon2020generating}). From the perspective of utilizing LLMs, methods have been proposed that combine with CoT techniques, allowing the model to self-generate subsequent questions and predict intermediate answers while performing multi-step reasoning (\citealp{yaoreact}; \citealp{khot2022decomposed}; \citealp{lazaridou2022internet}; \citealp{press-etal-2023-measuring}; \citealp{jiang-etal-2023-active}; \citealp{xu2024search}). However, these methods often involve complex processes where errors propagate and accumulate. To reduce this complexity, we aim to enhance LLMs' innate multi-hop reasoning capabilities.

In retrieval-augmented generation (\citealp{lewis2020retrieval}; \citealp{izacard2023atlas}; \citealp{shi2024replug}), the typical approach for multi-hop QA is to retrieve documents related to the query, and have the LLM reason based on this context in a "retrieve-and-reason" fashion. A key issue here is handling the noise introduced by irrelevant documents within the context. To address this, methods have been proposed to help the model correctly locate the evidence within the context (\citealp{shi-etal-2024-generate-then-ground}; \citealp{kimsure}; \citealp{yu2023chainofnote}; \citealp{yan2024corrective}). With the recent increase in the LLMs' context length, \citet{jiang2024longrag} have also explored including more documents and leveraging the LLM's reasoning capabilities. We aim to propose a method focusing on document order to enable the LLM to reason correctly within the context.

From the perspective of repetition in LLMs, \citet{springer2024repetition} demonstrate that by repeating sentences during the sentence embedding extraction process, it is possible to compute embeddings that encode bidirectional context even in unidirectional LMs. \citet{xu2023re} show that in arithmetic reasoning tasks, repeating the query can improve performance by leveraging bidirectional understanding. While these methods are similar to ours in that they employ repetition, we differ fundamentally in that we focus on the issue of document order in multi-hop reasoning rather than on simple bidirectional understanding.

From the perspective of context sensitivity in LLMs, recent works focus on addressing the issue of absolute positional bias in tasks where LLMs are required to choose between multiple options (\citealp{sensitivity1}; \citealp{sensitivity2}). Also, \citet{liu2023lostmiddlelanguagemodels} also address the issue of absolute positional bias, specifically the "lost in the middle" problem. This problem pertains to the model's difficulty in effectively recognizing information that appears in the middle positions of the context provided to the LLM. This absolute positional bias issue differs from the ordering problem we address in our work, which involves ensuring that LLMs interpret supporting documents within the context in the correct reasoning chain order, relative position, in the multi-hop reasoning task. \citet{sensitivity_math} analyze the sensitivity of reasoning performance in LLMs to the order in which rules or premises are presented during logical and mathematical reasoning. Our work investigates longer-context scenarios and addresses a broader spectrum of multi-hop reasoning tasks, including open-domain question answering and synthetic tasks, where document-level ordering issues are more pronounced. Furthermore, we propose a concrete methodology to mitigate order-related challenges, enhancing the novelty of our approach in addressing this problem.

\section{CoRe: Context Augmentation for Misordered Supporting Documents} \label{main_core}

In this section, we empirically demonstrate that the performance of LLMs in multi-hop reasoning is sensitive to the order of supporting documents within the context. We refer to this phenomenon as the \textbf{misordered context problem}. We theoretically formalize this problem from the perspective of context structure. Building on the formalization and considering its practical limitation, we propose \textbf{context repetition (CoRe)}. This method enables LLMs to interpret specific contiguous reasoning segments in an optimal order, facilitating more effective multi-hop reasoning within a given context.

\subsection{Problem of misordered Context in Multi-Hop Reasoning} \label{main_3_1}
We begin by addressing the following research question: \textit{Is the performance of LLMs on multi-hop reasoning tasks sensitive to the order of supporting documents within the context?} To explore this question, we conduct an empirical study in the context of the multi-hop QA task using the MuSiQue dataset. The dataset consists of samples categorized into 3 types: 2-hop, 3-hop, and 4-hop. For a given $k$-hop sample, there are $k$ supporting documents and several noisy documents that distract from the correct answer. To isolate the effect of supporting document order, we remove the noisy documents and construct clean contexts solely from the supporting documents for our experiments. 

For each $k$-hop query, we permute the order of the $k$ supporting documents in every possible way, resulting in a total of $k!$ different contexts. We then evaluate the model's response to each of these contexts. In other words, for each $k$-hop query, the model generates a total of $k!$ answers, and we assess these answers using the F1 score. We categorize the evaluation results based on the types of order of the documents within contexts associated with each query. The type is determined by the conditional probability of the correct answer $a$ for the model $\theta$, $p_\theta(a \mid Q, C, I)$ where $Q$ is the query, $C$ is the context in the specific order of the supporting documents, and $I$ is the instruction for question answering. The higher the probability, the more optimally the supporting documents are ordered within the context for the answering of the model. We evaluate the model's performance across the best-case (highest probability), the worst-case (lowest probability), and the mean performance across all possible contexts for each query.

Figure \ref{fig_dc_problem} presents the results of an empirical study on 2-hop, 3-hop, and 4-hop reasoning samples. For each $k$-hop type, we randomly sample 200 samples from the validation set for evaluation. We evaluate 4 different LLMs: Llama-3.1-8B-Instruct (\citealp{dubey2024llama3}), Mistral-7B-Instruct-v0.3 (\citealp{jiang2023mistral}), Qwen2.5-7B-Instruct (\citealp{qwen2.5}), and Phi-3.5-mini-instruct (\citealp{abdin2024phi}). Across all models, the performance gap between the best and worst cases is consistently pronounced, with at least 10 points in terms of F1 score. As the number of hops increases, indicating more complex reasoning types, the performance variation grows, reaching up to 26 points in terms of F1 score. Given that the context in this study consists solely of supporting documents, the performance disparity is expected to widen further in real-world tasks where noisy documents are included in the context.

Based on these findings, we demonstrate the answer to the research question: \textit{the multi-hop reasoning performance of LLMs is indeed significantly sensitive to the order in which supporting documents are presented}.
This also aligns with the rationale that decoder-only LLMs, operating with causal attention, interpret the context in a fixed left-to-right order.
In real-world scenarios, common prompting methods such as retrieval or search-based approaches do not guarantee that supporting documents will be presented to the model in an optimal order. Consequently, the model is often exposed to a misordered context that is disadvantageous for reasoning, leading to degraded performance. We term this issue the \textbf{misordered context} problem, and in the following sections, we describe our approach to addressing this problem.

\begin{figure*}[!t]
    \centering
    \includegraphics[width=\textwidth]{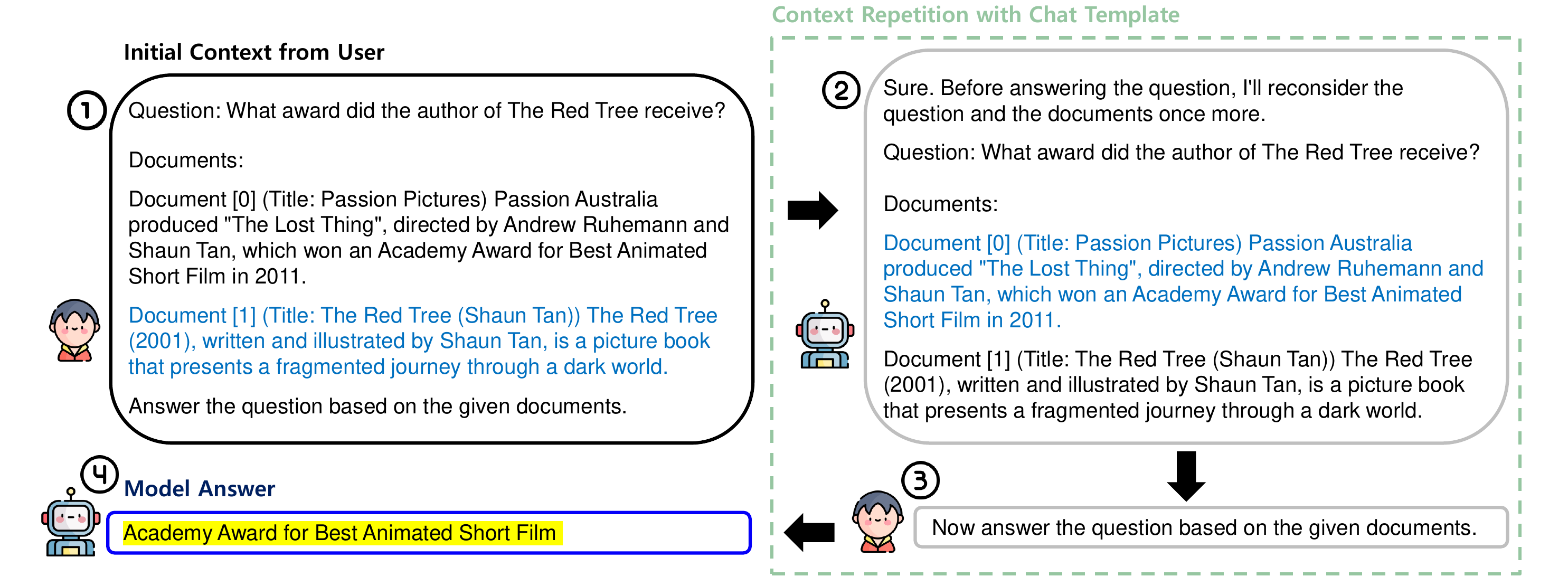} 
    \caption{
    Illustration of CoRe where the model understands the context with \textit{the optimal order of documents [1] and [0]}. Text with a white background is the prompt, and text with a yellow background is the model output.
    }
    \label{fig_method}
\end{figure*}
\subsection{Context Augmentation Approach}
For the misordered context problem, we formalize a solution from the perspective of context manipulation. We consider a $k$-hop reasoning scenario, where the model must derive an answer based on a prompt consisting of a query $Q$, context $C$, and instruction $I$. The context $C$ is composed of $k$ supporting documents as well as multiple noisy documents, structured as follows:

\begin{equation}\label{eq_context}
    C = (n_0, d_1, n_1, \cdots, d_k, n_k),
\end{equation}

\noindent where $d_i$ is the $i$-th supporting document within the context, and $n_i$ is an arbitrary number of noisy documents located between supporting documents, as well as at the beginning and the end of the context. At this point, we define an order $\sigma$ for the supporting documents as follows:

\begin{definition}\label{def_sigma}
    Let $\sigma(\cdot)$ be a permutation function for $(1, 2, \cdots, k)$. Then, order $\sigma$ is defined as
    \[
    \sigma(d_1, d_2, \cdots, d_k) \triangleq (d_{\sigma(1)}, d_{\sigma(2)}, \cdots, d_{\sigma(k)}).
    \]
\end{definition}

Based on the definition of order, we define the set of contexts that include supporting documents in a specific order $\sigma$ as follows:

\begin{definition}\label{def_sigma_context}
    Let $\sigma$ be an order of supporting documents. Then, we define $\mathcal{C}_\sigma$ as the set of all contexts that include the supporting documents in order $\sigma$:
    \[
    \mathcal{C}_{\sigma} \triangleq \left\{ 
    C \mid C = (n_0, d_{\sigma(1)}, n_1, \cdots, d_{\sigma(k)}, n_k), 
    \right. 
    \]
    \[
    \left. \forall n_0, n_1, \cdots, n_k
    \right\}.
    \]  
\end{definition}

The order $\sigma$ defined above can exist in a total of $k!$ different configurations, and thus there are $k!$ possible sets $\mathcal{C}_\sigma$. The initially provided context for answering the query belongs to some arbitrary $\mathcal{C}_\sigma$, and when $\sigma$ is the optimal order, the model is expected to perform at its best. We define the optimal order $\sigma^*_\theta$ for a specific model $\theta$ as follows:

\begin{definition}\label{def_optimal_sigma}
    Let $\theta$ be a model that performs multi-hop reasoning. Then, we define $\sigma^*_\theta$ as the optimal order of the given documents $d$ for the model $\theta$:
    \[
    \sigma^*_\theta \triangleq 
    \operatorname*{argmax}_{\sigma} \mathop{\mathbb{E}}_{C \in \mathcal{C}_\sigma} 
    \left[
        p_\theta\left (a \mid Q, C, I \right)
    \right].
    \]
\end{definition}

That is, the optimal order of supporting documents for a given model is defined as the order that maximizes the expected probability of generating the correct answer in the presence of arbitrary noisy documents. Based on the formalization so far, we aim to find a function that maps a given context $C$ to another context $C'$, which has the supporting documents in the optimal order. The goal is to increase the probability of the model generating the correct answer. Therefore, we introduce an augmentation function $f(\cdot)$ that satisfies the following conditions:

\begin{equation}\label{eq_augmentation}
    f(C) \in \mathcal{C}_{\sigma_\theta^*}.
\end{equation}

\subsection{Context Repetition} \label{3.3}
Now, we study the augmentation function discussed earlier. We define an augmentation function $f_{\text{rep}}^{(k)}(\cdot)$ as follows:

\begin{equation}\label{eq_rep_fn}
    f_{\text{rep}}^{(k)}(C) = \underbrace{C \oplus C \oplus \dots \oplus C}_{k \text{ times}}.
\end{equation}

Here, $C$ represents the given context, and $\oplus$ denotes the concatenation operation. Hence, the function $f_{\text{rep}}^{(k)}(C)$ creates an augmented context by just repeating the given context $k$ times.
Please note that $k=1$ indicates the absence of repetition.
The following theorem provides an important property of $f_{\text{rep}}^{(k)}(\cdot)$ in relation to $\sigma$:

\begin{theorem}\label{theorem}
     
    For a given context $C$ with its supporting documents ordered according to some permutation $\tau$, the augmented context $f_{\text{rep}}^{(k)}(C)$ belongs to the set $\mathcal{C}_\sigma$ for any permutation $\sigma$. Formally,
    \[
    f_{\text{rep}}^{(k)}(C) \in \mathcal{C}_\sigma, \quad \forall \sigma.
    \]
\end{theorem}

The proof of this theorem can be found in Appendix \ref{appendix_proofs}. From Theorem \ref{theorem}, we can derive the following corollary:

\begin{corollary}\label{corollary}
    $f_{\text{rep}}^{(k)}(C)$ always belongs to the set of all contexts that present the supporting documents in the optimal order. Formally, 
    \[
    f_{\text{rep}}^{(k)}(C) \in \mathcal{C}_{\sigma_\theta^*}.
    \]
\end{corollary}

Based on Corollary \ref{corollary}, we conclude that $f_{\text{rep}}^{(k)}(\cdot)$ is an appropriate augmentation function that satisfies Eq. \ref{eq_augmentation}. 

\subsection{Practical Approach}
From the previous section \ref{3.3}, we can conclude that $f_{\text{rep}}^{(k)}(\cdot)$ can serve as a methodology for mitigating the misordered context problem. However, for direct application, the following two issues must be addressed:
\begin{enumerate}[leftmargin=*, nolistsep]
    \item It is necessary to know the number of supporting documents in the given context, $k$. (Issue 1)
    \item The model should effectively handle the computational cost induced by the increased length when the context is repeated $k$ times. (Issue 2)
\end{enumerate}
In real-world multi-hop reasoning scenarios, it is not feasible to fully address issue 1. Additionally, issue 2 becomes intractable as the number of documents included in the context grows, and more importantly, it incurs a significant additional cost in terms of memory. 

To overcome these critical limitations, we replace $k$ with a predetermined hyperparameter $\hat{k}$, resulting the function $f_{\text{rep}}^{(\hat{k})}(\cdot)$ which can be considered an approximation of $f_{\text{rep}}^{(k)}(\cdot)$. This directly addresses the issue 1 by constraining the unknown value of $k$ to predetermined $\hat{k}$. While this approach does not guarantee that all supporting documents will be recognized in the optimal order, it ensures that \textit{at least the $\hat{k}$-contiguous reasoning chains within supporting documents are interpreted in the optimal order}. This allows the LLM to perform multi-hop reasoning from the optimal starting point, guiding the model to execute subsequent reasoning steps correctly.  From the perspective of the issue 2, the introduction of $\hat{k}$ significantly reduces the additional computational and memory costs, bringing them to a manageable level. 

Consequently, we propose to use $f_{\text{rep}}^{(\hat{k})}(\cdot)$ as a method for solving the misordered context problem. In other words, we introduce the \textbf{co}ntext \textbf{re}petition (\textbf{CoRe}) methodology, which repeatedly presents the given context to guide the model recognize the supporting documents in the appropriate order. Figure \ref{fig_method} illustrates the detailed operation of our methodology when $\hat{k}=2$. In practice, we implement this method by utilizing a chat template to ensure that the model naturally repeats the context given by the user.

\section{Experiments} \label{main_experiments}
In this section, we evaluate CoRe on multi-hop QA tasks and a synthetic task that requires complex reasoning. In addition to the main experiments, we conduct an analysis of the mechanism through which CoRe achieves performance improvements in the model.

\subsection{Experimental Settings}
We conduct experiments on three multi-hop QA benchmarks: HotpotQA, 2WikiMultihopQA, and MuSiQue. For HotpotQA and 2WikiMultihopQA, reasoning-type annotations are provided for each query. For these two datasets, we create evaluation sets by sampling an equal number of queries from each reasoning type within the validation set. In the case of MuSiQue, as described in Section \ref{main_3_1}, query types are categorized by the number of hops. We use the entire validation set as the evaluation set for this dataset. To construct the context of multiple documents, we combine the supporting documents with distracting (noisy) documents assigned to each sample in the datasets. Detailed statistics for each evaluation set are provided in Appendix \ref{appendix_data_statics}. In the multi-hop QA evaluation, we set the $\hat{k}$ within the range of $[1,3]$ ($\hat{k}=1$ means the naive LLM baseline without repetition). 

In addition to the multi-hop QA tasks, we evaluate model performance on an additional synthetic task.
Specifically, we provide the model information about a list that contains non-negative integers as elements in a chained form consisting of two consecutive elements. Afterward, we ask the model for the first element of the list that contains a specific element. At this point, the specific element is the last element of a list. For example, when there are two lists with 3 elements each, the context and question are structured as follows:



\begin{lstlisting}
All the 2 lists described below contain exactly 3 elements.
In the list 0, 381 is positioned immediately before 512.
In the list 1, 7123 is positioned immediately before 34.
In the list 0, 512 is positioned immediately before 1021.
In the list 1, 34 is positioned immediately before 6397.
Question: What is the first element of the list that contains 1021?
\end{lstlisting}


Since the elements of each list are presented in sequential order from the first to the last, the process of finding the list that contains the last element and then searching for the first element of that list must be performed in a right-to-left order. This poses a challenge for models that typically process context in a left-to-right order. Moreover, since this type of task does not involve specific factual knowledge, it minimizes the interference of the model's pre-existing factual knowledge (\citealp{xiong2024artificial}). 
Thus, these characteristics of the synthetic task enable us to establish an environment that clearly exposes the model's inherent misordered context problem.
We create a synthetic dataset consisting of 1000 samples for the evaluation. Each sample is composed of 10 lists, each containing 3 integers between 0 and 9999 that do not overlap.

\begin{table*}[!t]
    \centering
    \resizebox{\textwidth}{!}{
        \renewcommand{\arraystretch}{1.1}
        \begin{tabular}{l||cc|cccc|ccc}
            \toprule
            \multicolumn{1}{c||}{\multirow{2}{*}{\textbf{Models}}} & \multicolumn{2}{c|}{\textbf{HotpotQA}} & \multicolumn{4}{c|}{\textbf{2WikiMultihopQA}} & \multicolumn{3}{c}{\textbf{MuSiQue}}                                                                                                          \\
            \multicolumn{1}{c||}{}                                 & Comparison                             & Bridge                                        & Compositional                        & Comparison     & Inference      & Bridge-Comparison & 2-hop          & 3-hop          & 4-hop          \\ \hline
            Llama-3.1-8B                                           & 49.31                         & 61.19                                         & 38.10                                & 55.77          & 40.75          & 34.86             & 32.05          & 26.24          & 27.06          \\
            + CoRe ($\hat{k}=2$)                                                & 46.93                                  & 66.82                                & 48.15                       & 58.13 & 44.81 & 65.11    & 39.10 & 38.23 &34.08\\ 
            + CoRe ($\hat{k}=3$)                                                & \textbf{50.46}& \textbf{70.54}& \textbf{50.74}& \textbf{62.58}& \textbf{47.44}& \textbf{68.18}& \textbf{43.64}& \textbf{41.02}& \textbf{36.08}\\ \hline
            Mistral-7B                                             & 34.42                         & 41.14                                         & 18.82                                & 32.90          & 29.77          & 15.43             & 15.14          & 10.31          & 11.00          \\
            + CoRe ($\hat{k}=2$)                                                 & 30.41                                  & 45.79                                & 27.78                       & 35.67 & 36.39 & \textbf{29.09}    & 20.6  & 15.74 & 12.36 \\ 
            + CoRe ($\hat{k}=3$)                                                & \textbf{42.50}&\textbf{ 54.51}& \textbf{38.93}& \textbf{42.91}& \textbf{38.67}& 28.70& \textbf{23.96}& \textbf{20.17}& \textbf{14.78}\\ \hline
            Qwen2.5-7B                                             & 61.96                                  & 61.66                                         & 41.36                                & 65.47          & 37.26          & 39.45             & 33.56          & 34.87          & 37.49          \\
            + CoRe ($\hat{k}=2$)                                                 & \textbf{67.03}                         & 70.33                                & \textbf{54.20}                       & \textbf{70.30} & 42.40 & 64.54    & 44.76 & \textbf{43.10}  & 45.08 \\
            + CoRe ($\hat{k}=3$)                                                & 63.37& \textbf{71.59}& 54.13& 69.64& \textbf{44.30}& \textbf{68.00}& \textbf{46.12}& 42.84& \textbf{45.88}\\ \hline
            Phi-3.5-mini                                           & 26.44                         & 44.17                                         & 27.70                                & 45.26          & 27.00          & 35.19             & 17.60          & 16.80          & 12.67          \\
            + CoRe ($\hat{k}=2$)                                                  & 23.29                                  & 49.67                                & 37.64                       & \textbf{50.92} & 32.34 & \textbf{36.71}    & 28.60 & 26.31 & 26.59 \\ 
            + CoRe ($\hat{k}=3$)                                                & \textbf{41.41}& \textbf{59.07}& \textbf{40.62}& 46.91& \textbf{33.76}& 22.98& \textbf{31.49}& \textbf{26.83}& \textbf{32.40}\\ \hline
            GPT-4o-mini                                            & \textbf{66.92}                         & 68.66                                         & 51.96                                & 76.63          & 43.84          & 60.96             & 45.26          & 42.04          & 40.04          \\
            + CoRe ($\hat{k}=2$)                                                  & 64.72                                  & \textbf{72.67  }                              & \textbf{57.43}                       & 79.14 & \textbf{52.03} & 66.96    & 52.47 & \textbf{49.47} & 42.53 \\
            + CoRe ($\hat{k}=3$)                                                & 66.52& \textbf{74.06}& 56.72& \textbf{79.84}& 51.09& \textbf{68.84}& \textbf{52.66}& 49.37& \textbf{42.89}\\ 
            \bottomrule
        \end{tabular}
    }
    \caption{Main results of F1 score in the multi-hop QA tasks. All models are instruction-tuned LLMs.}
    \label{tab_main}
\end{table*}

We conduct experiments using a total of five models, including GPT-4o-mini (\citealp{gpt4omini}), in addition to the four LLMs used in section \ref{main_3_1}. For Multi-hop QA tasks, we evaluate by measuring the F1 score based on the short answers generated by the models. We evaluate synthetic tasks by measuring accuracy based on whether the answers generated by the models match the correct answers. All prompts used in the experiments can be found in the Appendix \ref{appendix_prompts}.

\begin{figure}[t]
{
\begin{center}
\centerline{\includegraphics[width=\columnwidth]{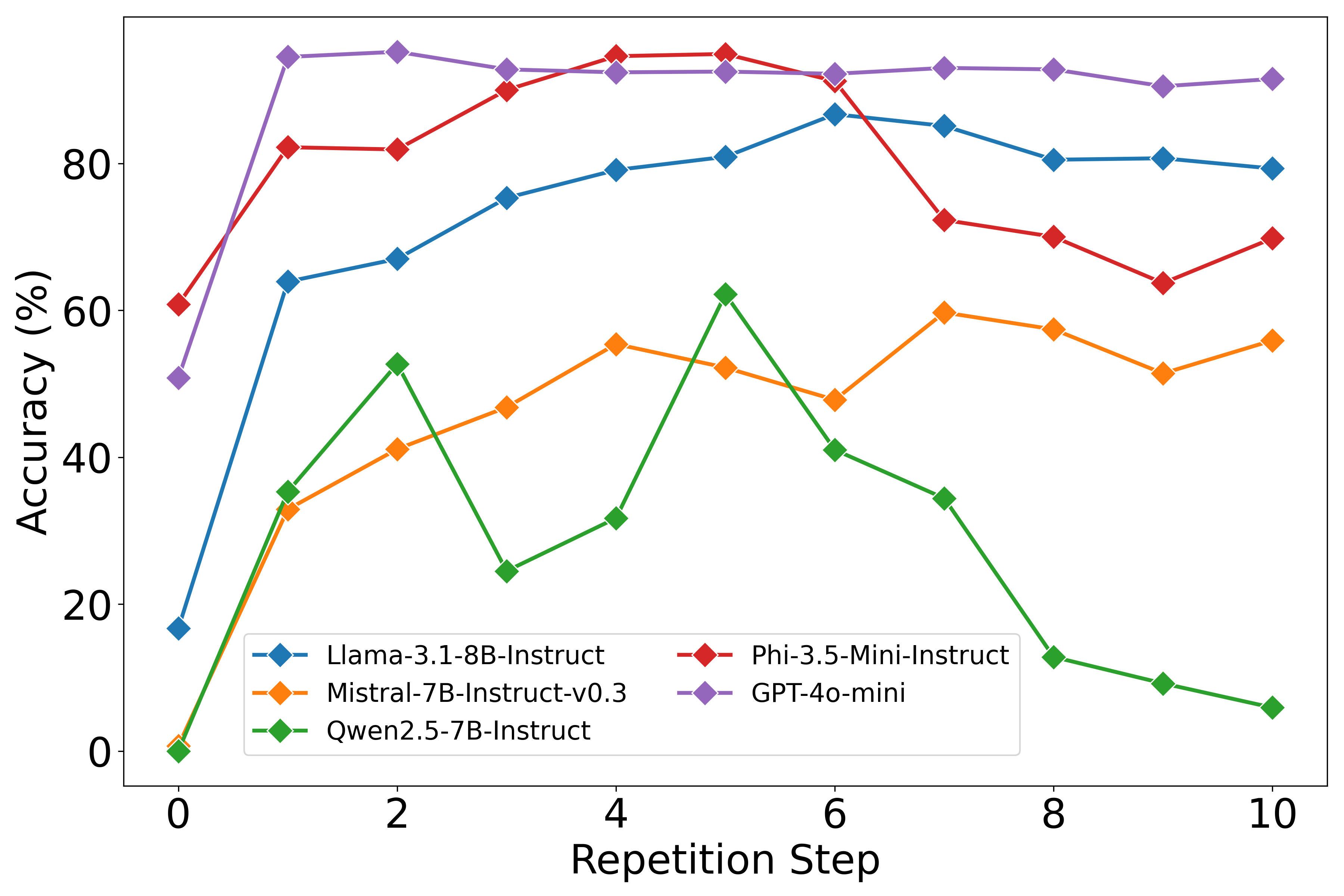}}
\caption{Main results in the synthetic task. Repetition step denotes the number of additional repetitions ($\hat{k}-1$).}
\label{fig_main_synthetic}
\end{center}
}
\end{figure}

\subsection{Main Results}
Table \ref{tab_main} shows the F1 scores of baseline models and models with CoRe applied for multi-hop QA tasks. Overall, performance improvements were observed across all models and datasets. 
Notably, the performance appears to reach a plateau in the $\hat{k}=2$ setting, where the context is repeated only once. Furthermore, we find additional performance improvements when further repetitions are applied ($\hat{k}=3$).
In the case of HotpotQA, a significant performance improvement was seen in the bridge type. This is because answering bridge-type questions requires reasoning across documents by linking information between them, which is closely related to the order of documents. For 2WikiMultihopQA, a considerable performance improvement was observed in the bridge-comparison type, with Llama achieving a 30-point improvement and Qwen achieving a 25-point improvement in the F1 score. This is because the bridge-comparison type is a complex 4-hop reasoning task that combines bridge and comparison from HotpotQA. The MuSiQue dataset's overall performance improved regardless of the number of hops. These results indicate that CoRe substantially enhances the model's multi-hop reasoning performance, particularly in tasks where the order of supporting documents is crucial, effectively guiding the model along the correct reasoning paths.
\begin{figure*}[!t]
    \centering
    \includegraphics[width=\textwidth]{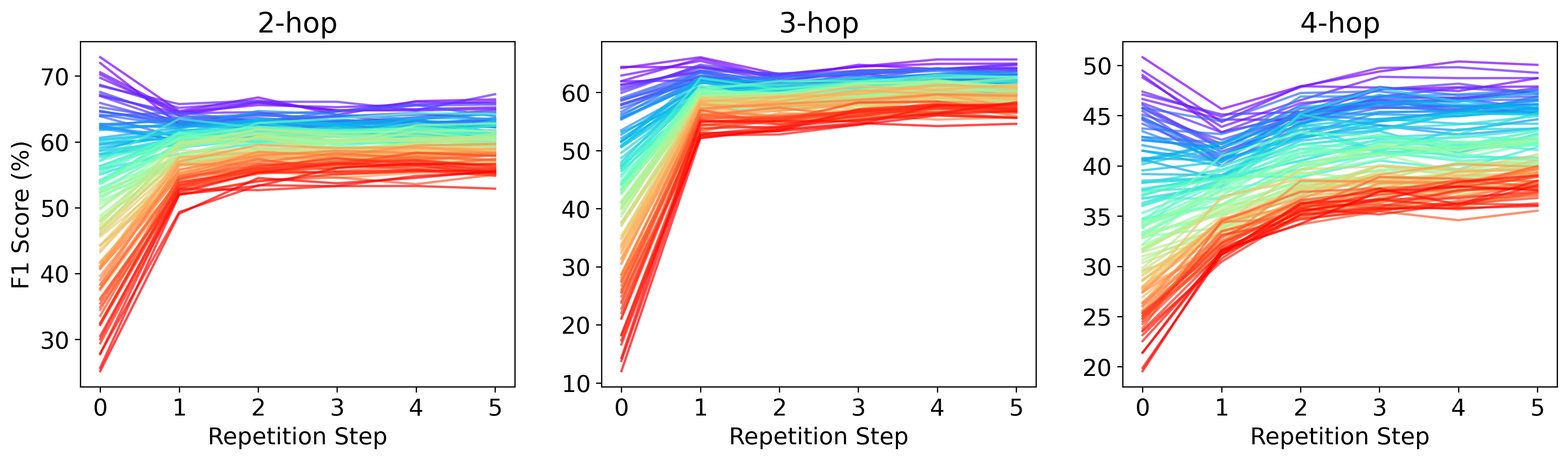} 
    \caption{
    Performance of Llama-3.1-8B-Instruct with permuted contexts of MuSiQue during repetitions. The red line denotes the context in the worst order, and the purple line denotes the context in the best order.
    }
    \label{fig_analysis_permute_llama}
\end{figure*}
\begin{figure}[t]
{
\begin{center}
\centerline{\includegraphics[width=\columnwidth]{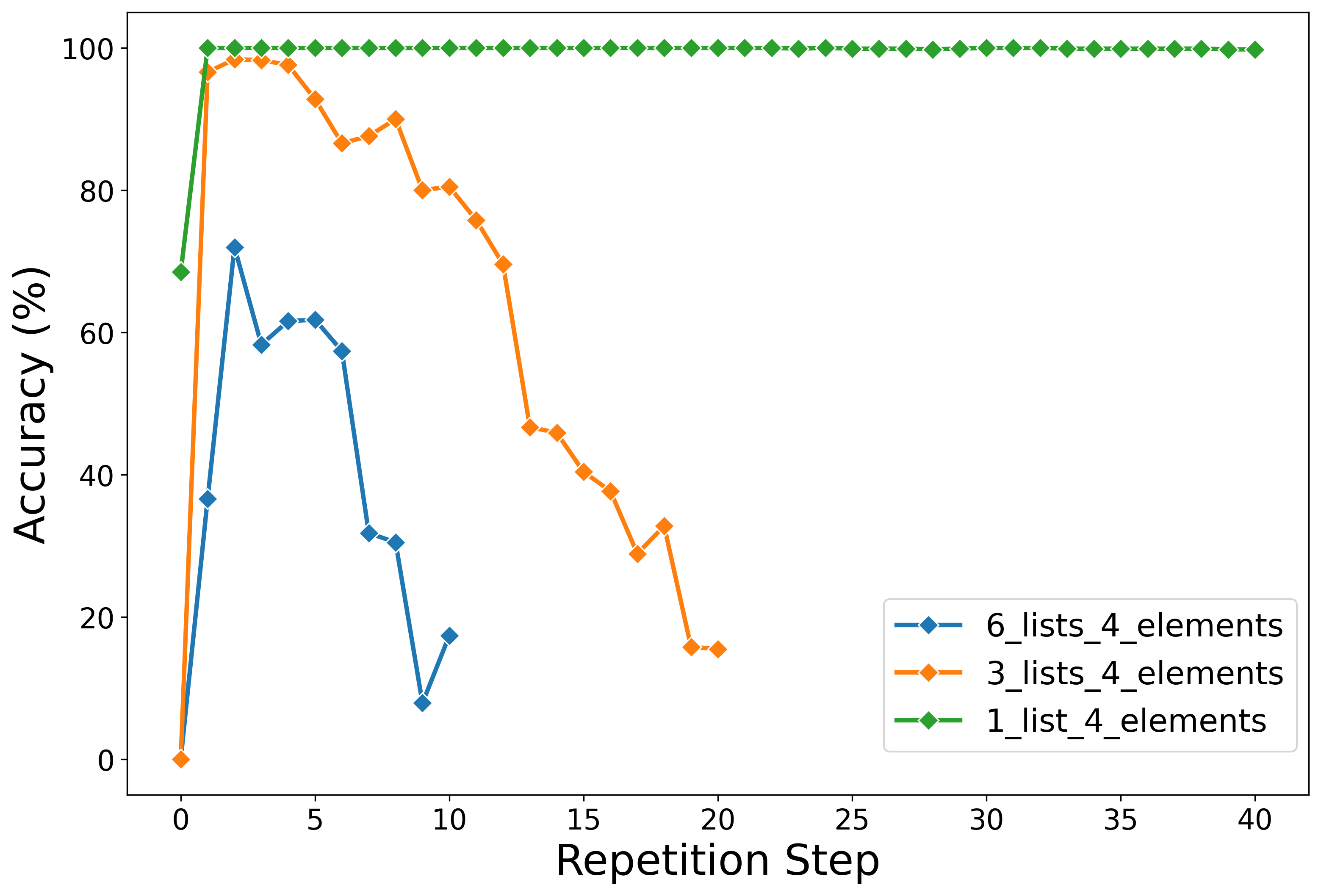}}
\caption{Performance of Qwen2.5-7B-Instruct for contexts with various noise powers in the synthetic task.}
\label{fig_analaysis_synthetic}
\end{center}
}
\end{figure}

We measure the accuracy of each model as we increase the number of context repetitions up to 10 ($\hat{k}=11$) in the synthetic task. As shown in Figure \ref{fig_main_synthetic}, we can see that the accuracy when no repetition is performed has a low value between 0\% and 60\%. This demonstrates that existing LLMs are unable to perform the synthetic task directly at once. However, when CoRe is applied, we observe a significant performance improvement, with accuracy increasing by at least 30\% points and up to 70\% points. Additionally, we can see that many models tend to improve performance as the number of repetitions increases. This suggests that the model leads to the more appropriate order of reasoning chains as repetition progresses multiple times. However, for some models, particularly Qwen, there is a tendency for performance to drop again as the repetition steps increase. This appears to be due to noisy information being repeated in the context, negatively affecting its reasoning ability. We provide a detailed analysis of this in Section \ref{sec_4_3}.

\subsection{Analysis} \label{sec_4_3}
\paragraph{Does CoRe really guide the model along the better reasoning chain?}
We conduct additional analysis to investigate whether the performance improvement from CoRe is indeed related to the order of the documents. Similar to the experiments described in Section \ref{main_3_1}, we permute the order of the documents in the context across all possible combinations and measure the model's performance for each permuted context. Specifically, for 200 $k$-hop samples from MuSiQue, we construct a context with five documents, including the $k$ supporting documents and some random $5-k$ noisy documents assigned to each sample. As a result, each sample has 120 permuted contexts, and we sort the performance of each context from the worst order to the best order, based on $p_\theta(a \mid Q, C, I)$. Figure \ref{fig_analysis_permute_llama} shows the results for Llama.
We plot the worst order with a red line and the best with a purple line. For other cases, we plot their performance with the line of a continuous gradient of colors from red to purple, corresponding to the performance range from the worst to the best case.
In other words, the closer the color is to red in the rainbow spectrum, the lower $p_\theta(a \mid Q, C, I)$ is, and the closer to purple, the higher $p_\theta(a \mid Q, C, I)$ is.
Across all query types, we find compelling evidence that CoRe contributes to performance improvements in terms of document order: \textbf{The closer the context is to the worst order case, the greater the performance improvement from repetition}. Additionally, we observe that as the number of hops increases, performance improves progressively with iterative repetition, indicating that multiple repetitions are more effective when the required reasoning steps increase. Appendix \ref{appendix_addi_results} shows the results for other LLMs in this analysis.

\paragraph{Impact of noisy documents}
We further analyze the impact of noisy documents on CoRe. In the synthetic task, we measure accuracy by increasing the number of repetitions in cases where the number of lists is 6, 3, and 1, respectively. Here, the decrease in the number of lists means a reduction in the proportion of information unrelated to the search target. Figure \ref{fig_analaysis_synthetic} shows the results of this analysis. Considering that the length of the context decreases as the number of lists handled in the context decreases, we perform more repetitions in cases with fewer lists. As a result, we confirm that as the proportion of noisy information within the context decreases, the performance degradation caused by excessive repetition is reduced. In particular, when there is no noisy information, performance did not decrease even with a large number of repetitions. Since noisy information adversely affects model reasoning at high repetition counts, exploring the valid number of repetitions required for model reasoning could be an interesting future work based on our findings.



\section{Further Studies} \label{main_further_studies}

\subsection{Robustness to Positional Bias}

\begin{figure}[t]
{
\begin{center}
\centerline{\includegraphics[width=\columnwidth]{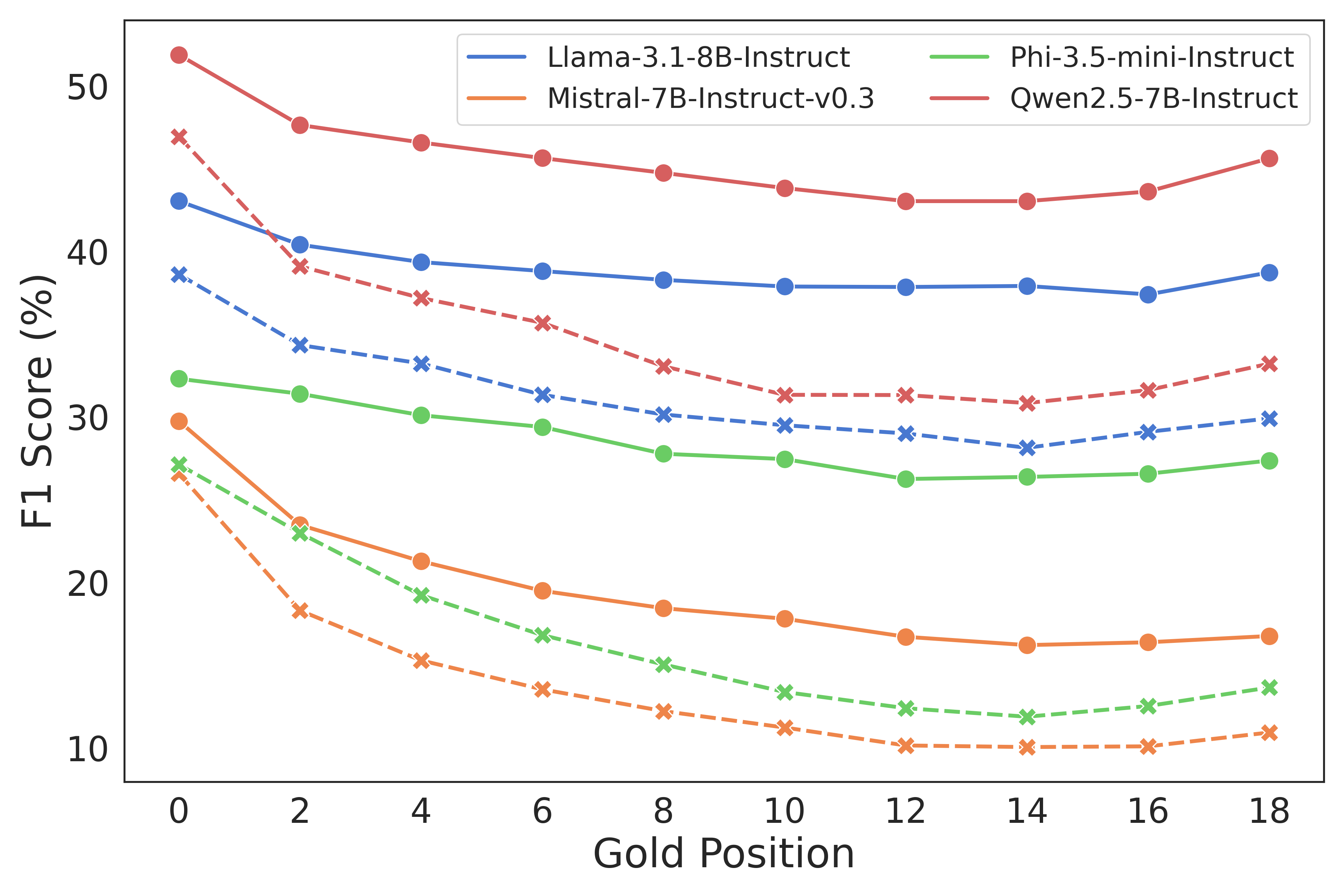}}
\caption{Performance of LLMs for various positions of the supporting document. The solid line represents CoRe, and the dotted line represents the baseline.}
\label{fig_robust_position}
\end{center}
}
\end{figure}

We analyze the absolute positional bias in the context based on the position of supporting documents. Here, positional bias refers to the phenomenon where the QA performance of an LLM varies depending on the position of the supporting documents within the context, known as the "lost-in-the-middle" problem \cite{liu2023lostmiddlelanguagemodels}. We concatenate $k$ supporting documents in MuSiQue and measure the F1 scores of the baseline without repetition ($\hat{k}=1$) and CoRe ($\hat{k}=2$) while shifting the position of the documents from 0 to 18 by increments of 2. As shown in \ref{fig_robust_position}, we observe that the performance gap between the baseline and CoRe increases as the position of the supporting documents moves toward the middle. Particularly for Llama (blue line) and Phi (green line), while the baseline suffers from the "lost-in-the-middle" problem, CoRe demonstrates robustness against positional bias, exhibiting minimal performance variation depending on the position of the supporting documents. Based on these results, we demonstrate that CoRe is a solution to mitigate the "lost-in-the-middle" problem.

\subsection{Enhancement in Retrieve-and-Reason} \label{sec_5_2}
We evaluate the effectiveness of CoRe ($\hat{k}=2$) in a retrieve-and-reason task, which closely mirrors real-world scenarios. 
We utilize Contriever (\citealp{izacard2021contriever}) as the retriever model, constructing the context with up to 100 retrieved documents from MuSiQue samples. We use Llama for this experiment. We first assess CoRe's performance with (1) the standard RAG, which directly retrieves the top 100 relevant documents to form the context. We also evaluate CoRe with two additional methods: (2) IRCoT (\citealp{trivedi-etal-2023-interleaving}), where retrieval is performed sequentially during the CoT reasoning process, and (3) Decompose, where the query is decomposed into sub-queries, and reasoning is conducted by deriving intermediate answers for each sub-query (\citealp{press-etal-2023-measuring}). As shown in Table \ref{tab_retrieve_reason}, CoRe significantly improves performance across all methods, demonstrating its effectiveness when integrated into various retrieval-based reasoning approaches. Appendix \ref{appendix_detail_retrieve} presents the experimental details of this section.

\begin{table}[!t]
    \centering
    \resizebox{0.8\columnwidth}{!}{
        \renewcommand{\arraystretch}{1.1}
        \begin{tabular}{l|ccc}
            \toprule
            \multicolumn{1}{c|}{\textbf{Methods}} & \textbf{2-hop} & \textbf{3-hop} & \textbf{4-hop} \\\hline
            RAG                                   & 12.88          & 8.06           & 6.70           \\
            + CoRe ($\hat{k}=2$)                               & \textbf{17.04} & \textbf{12.44} & \textbf{9.07}  \\ \hline
            IRCoT                                & 23.67          & 10.64          & 4.33           \\
            + Dec.                                & 29.03          & 11.78          & 6.62           \\
            + CoRe ($\hat{k}=2$)                              & \textbf{37.65} & \textbf{19.35} & 8.90           \\
            + CoRe \& Dec.                        & 36.62          & 17.87          & \textbf{11.52} \\
            \bottomrule
        \end{tabular}
    }
    \caption{F1 results for retrieve-and-reason tasks in MuSiQue. Dec. refers to Decompose method.}
    \label{tab_retrieve_reason}
\end{table}

Additionally, we report the performance of CoRe when applied in Chain-of-Thought (CoT) scenarios in Appendix \ref{appendix_addi_results}. 
We also conduct various ablation studies in Appendix \ref{appendix_ablations}.

\section{Conclusion} \label{main_conclusion}
In this work, we introduce the misordered context problem, which critically impacts LLMs' multi-hop reasoning capabilities. Formalizing this problem regarding context augmentation, we propose a simple yet highly effective solution: the CoRe method, which optimally arranges contiguous reasoning segments within the context to enhance performance. We demonstrate the contributions of our approach across various multi-hop reasoning benchmarks, including synthetic tasks, and elucidate its effectiveness through further studies. Our work highlights a key issue crucial for future discussions in multi-hop reasoning tasks with LLMs. Furthermore, we believe CoRe could provide meaningful insights for research on the test-time scaling of LLMs that perform complex reasoning.

\section*{Limitations}
The context repetition method proposed in this work increases the input context length by repeating the prompt's contents, which introduces additional costs for memory and time during inference. While the time cost is kept manageable due to the efficient KV caching mechanism, future research could explore ways to repeat only the core contents of the context, thereby minimizing unnecessary expansion. Moreover, as discussed in the analysis section, it is crucial to address the performance degradation caused by noisy documents when repeating the context. A promising avenue for future work would be to combine the inherent capabilities of LLMs with techniques that minimize the influence of noisy documents while maximizing the effectiveness of context repetition. Overall, our work lays the foundational groundwork for addressing the misordered context problem through context augmentation, and future studies could build upon this foundation to further refine and extend the approach.

\section*{Acknowledgements}
This work was supported by the National Research Foundation of Korea (NRF) grant funded by the Korea government (MSIT) (No. 2022R1A3B1077720), 
Institute of Information \& communications Technology Planning \& Evaluation (IITP) grant funded by the Korea government(MSIT) [NO.RS-2021-II211343, Artificial Intelligence Graduate School Program (Seoul National University)],
the BK21 FOUR program of the Education and Research Program for Future ICT Pioneers, Seoul National University, 
Samsung Electronics Co., Ltd (IO240311-09242-01), 
and a grant from the Yang Young Foundation.

\bibliography{custom}

\begin{thebibliography}{47}
\providecommand{\natexlab}[1]{#1}

\bibitem[{Abdin et~al.(2024)Abdin, Jacobs, Awan, Aneja, Awadallah, Awadalla, Bach, Bahree, Bakhtiari, Behl et~al.}]{abdin2024phi}
Marah Abdin, Sam~Ade Jacobs, Ammar~Ahmad Awan, Jyoti Aneja, Ahmed Awadallah, Hany Awadalla, Nguyen Bach, Amit Bahree, Arash Bakhtiari, Harkirat Behl, et~al. 2024.
\newblock Phi-3 technical report: A highly capable language model locally on your phone.
\newblock \emph{arXiv preprint arXiv:2404.14219}.

\bibitem[{Brown et~al.(2020)Brown, Mann, Ryder, Subbiah, Kaplan, Dhariwal, Neelakantan, Shyam, Sastry, Askell et~al.}]{gpt3}
Tom~B Brown, Benjamin Mann, Nick Ryder, Melanie Subbiah, Jared Kaplan, Prafulla Dhariwal, Arvind Neelakantan, Pranav Shyam, Girish Sastry, Amanda Askell, et~al. 2020.
\newblock Language models are few-shot learners.
\newblock In \emph{Proceedings of the 34th International Conference on Neural Information Processing Systems}, pages 1877--1901.

\bibitem[{Chen et~al.(2024)Chen, Chi, Wang, and Zhou}]{sensitivity_math}
Xinyun Chen, Ryan~Andrew Chi, Xuezhi Wang, and Denny Zhou. 2024.
\newblock Premise order matters in reasoning with large language models.
\newblock In \emph{Forty-first International Conference on Machine Learning}.

\bibitem[{Cuconasu et~al.(2024)Cuconasu, Trappolini, Siciliano, Filice, Campagnano, Maarek, Tonellotto, and Silvestri}]{Cuconasu_2024_powerofnoise}
Florin Cuconasu, Giovanni Trappolini, Federico Siciliano, Simone Filice, Cesare Campagnano, Yoelle Maarek, Nicola Tonellotto, and Fabrizio Silvestri. 2024.
\newblock \href {https://doi.org/10.1145/3626772.3657834} {The power of noise: Redefining retrieval for rag systems}.
\newblock In \emph{Proceedings of the 47th International ACM SIGIR Conference on Research and Development in Information Retrieval}, SIGIR 2024. ACM.

\bibitem[{Dubey et~al.(2024)Dubey, Jauhri, Pandey, Kadian, Al-Dahle, Letman, Mathur, Schelten, Yang, Fan et~al.}]{dubey2024llama3}
Abhimanyu Dubey, Abhinav Jauhri, Abhinav Pandey, Abhishek Kadian, Ahmad Al-Dahle, Aiesha Letman, Akhil Mathur, Alan Schelten, Amy Yang, Angela Fan, et~al. 2024.
\newblock The llama 3 herd of models.
\newblock \emph{arXiv preprint arXiv:2407.21783}.

\bibitem[{Gao et~al.(2023)Gao, Xiong, Gao, Jia, Pan, Bi, Dai, Sun, and Wang}]{gao2023retrievalsurvey}
Yunfan Gao, Yun Xiong, Xinyu Gao, Kangxiang Jia, Jinliu Pan, Yuxi Bi, Yi~Dai, Jiawei Sun, and Haofen Wang. 2023.
\newblock Retrieval-augmented generation for large language models: A survey.
\newblock \emph{arXiv preprint arXiv:2312.10997}.

\bibitem[{Ho et~al.(2020)Ho, Duong~Nguyen, Sugawara, and Aizawa}]{xanh2020_2wikimultihop}
Xanh Ho, Anh-Khoa Duong~Nguyen, Saku Sugawara, and Akiko Aizawa. 2020.
\newblock \href {https://www.aclweb.org/anthology/2020.coling-main.580} {Constructing a multi-hop {QA} dataset for comprehensive evaluation of reasoning steps}.
\newblock In \emph{Proceedings of the 28th International Conference on Computational Linguistics}, pages 6609--6625, Barcelona, Spain (Online). International Committee on Computational Linguistics.

\bibitem[{Izacard et~al.(2021)Izacard, Caron, Hosseini, Riedel, Bojanowski, Joulin, and Grave}]{izacard2021contriever}
Gautier Izacard, Mathilde Caron, Lucas Hosseini, Sebastian Riedel, Piotr Bojanowski, Armand Joulin, and Edouard Grave. 2021.
\newblock \href {https://doi.org/10.48550/ARXIV.2112.09118} {Unsupervised dense information retrieval with contrastive learning}.

\bibitem[{Izacard et~al.(2023)Izacard, Lewis, Lomeli, Hosseini, Petroni, Schick, Dwivedi-Yu, Joulin, Riedel, and Grave}]{izacard2023atlas}
Gautier Izacard, Patrick Lewis, Maria Lomeli, Lucas Hosseini, Fabio Petroni, Timo Schick, Jane Dwivedi-Yu, Armand Joulin, Sebastian Riedel, and Edouard Grave. 2023.
\newblock Atlas: Few-shot learning with retrieval augmented language models.
\newblock \emph{Journal of Machine Learning Research}, 24(251):1--43.

\bibitem[{Jiang et~al.(2023{\natexlab{a}})Jiang, Sablayrolles, Mensch, Bamford, Chaplot, Casas, Bressand, Lengyel, Lample, Saulnier et~al.}]{jiang2023mistral}
Albert~Q Jiang, Alexandre Sablayrolles, Arthur Mensch, Chris Bamford, Devendra~Singh Chaplot, Diego de~las Casas, Florian Bressand, Gianna Lengyel, Guillaume Lample, Lucile Saulnier, et~al. 2023{\natexlab{a}}.
\newblock Mistral 7b.
\newblock \emph{arXiv preprint arXiv:2310.06825}.

\bibitem[{Jiang et~al.(2023{\natexlab{b}})Jiang, Xu, Gao, Sun, Liu, Dwivedi-Yu, Yang, Callan, and Neubig}]{jiang-etal-2023-active}
Zhengbao Jiang, Frank Xu, Luyu Gao, Zhiqing Sun, Qian Liu, Jane Dwivedi-Yu, Yiming Yang, Jamie Callan, and Graham Neubig. 2023{\natexlab{b}}.
\newblock \href {https://doi.org/10.18653/v1/2023.emnlp-main.495} {Active retrieval augmented generation}.
\newblock In \emph{Proceedings of the 2023 Conference on Empirical Methods in Natural Language Processing}, pages 7969--7992, Singapore. Association for Computational Linguistics.

\bibitem[{Khot et~al.(2022)Khot, Trivedi, Finlayson, Fu, Richardson, Clark, and Sabharwal}]{khot2022decomposed}
Tushar Khot, Harsh Trivedi, Matthew Finlayson, Yao Fu, Kyle Richardson, Peter Clark, and Ashish Sabharwal. 2022.
\newblock Decomposed prompting: A modular approach for solving complex tasks.
\newblock \emph{arXiv preprint arXiv:2210.02406}.

\bibitem[{Kim et~al.(2024)Kim, Nam, Mo, Park, Lee, Seo, Ha, and Shin}]{kimsure}
Jaehyung Kim, Jaehyun Nam, Sangwoo Mo, Jongjin Park, Sang-Woo Lee, Minjoon Seo, Jung-Woo Ha, and Jinwoo Shin. 2024.
\newblock Sure: Summarizing retrievals using answer candidates for open-domain qa of llms.
\newblock In \emph{The Twelfth International Conference on Learning Representations}.

\bibitem[{Kojima et~al.(2022)Kojima, Gu, Reid, Matsuo, and Iwasawa}]{kojima2022zeroshotcot}
Takeshi Kojima, Shixiang~Shane Gu, Machel Reid, Yutaka Matsuo, and Yusuke Iwasawa. 2022.
\newblock Large language models are zero-shot reasoners.
\newblock \emph{Advances in neural information processing systems}, 35:22199--22213.

\bibitem[{Lazaridou et~al.(2022)Lazaridou, Gribovskaya, Stokowiec, and Grigorev}]{lazaridou2022internet}
Angeliki Lazaridou, Elena Gribovskaya, Wojciech Stokowiec, and Nikolai Grigorev. 2022.
\newblock Internet-augmented language models through few-shot prompting for open-domain question answering.
\newblock \emph{arXiv preprint arXiv:2203.05115}.

\bibitem[{Lee et~al.(2024)Lee, Chen, Dai, Dua, Sachan, Boratko, Luan, Arnold, Perot, Dalmia et~al.}]{lee2024canlongcontext}
Jinhyuk Lee, Anthony Chen, Zhuyun Dai, Dheeru Dua, Devendra~Singh Sachan, Michael Boratko, Yi~Luan, S{\'e}bastien~MR Arnold, Vincent Perot, Siddharth Dalmia, et~al. 2024.
\newblock Can long-context language models subsume retrieval, rag, sql, and more?
\newblock \emph{arXiv preprint arXiv:2406.13121}.

\bibitem[{Lewis et~al.(2020)Lewis, Perez, Piktus, Petroni, Karpukhin, Goyal, K{\"u}ttler, Lewis, Yih, Rockt{\"a}schel et~al.}]{lewis2020retrieval}
Patrick Lewis, Ethan Perez, Aleksandra Piktus, Fabio Petroni, Vladimir Karpukhin, Naman Goyal, Heinrich K{\"u}ttler, Mike Lewis, Wen-tau Yih, Tim Rockt{\"a}schel, et~al. 2020.
\newblock Retrieval-augmented generation for knowledge-intensive nlp tasks.
\newblock \emph{Advances in Neural Information Processing Systems}, 33:9459--9474.

\bibitem[{Li et~al.(2024)Li, Liang, Lyu, and Wang}]{li2024makinglongcoxtllmbettermultihop}
Yanyang Li, Shuo Liang, Michael Lyu, and Liwei Wang. 2024.
\newblock Making long-context language models better multi-hop reasoners.
\newblock In \emph{Proceedings of the 62nd Annual Meeting of the Association for Computational Linguistics (Volume 1: Long Papers)}, pages 2462--2475.

\bibitem[{Liu et~al.(2023)Liu, Lin, Hewitt, Paranjape, Bevilacqua, Petroni, and Liang}]{liu2023lostmiddlelanguagemodels}
Nelson~F. Liu, Kevin Lin, John Hewitt, Ashwin Paranjape, Michele Bevilacqua, Fabio Petroni, and Percy Liang. 2023.
\newblock \href {https://arxiv.org/abs/2307.03172} {Lost in the middle: How language models use long contexts}.
\newblock \emph{Preprint}, arXiv:2307.03172.

\bibitem[{Malon and Bai(2020)}]{malon2020generating}
Christopher Malon and Bing Bai. 2020.
\newblock Generating followup questions for interpretable multi-hop question answering.
\newblock \emph{arXiv preprint arXiv:2002.12344}.

\bibitem[{Mavi et~al.(2024)Mavi, Jangra, Jatowt et~al.}]{mavi2024multi}
Vaibhav Mavi, Anubhav Jangra, Adam Jatowt, et~al. 2024.
\newblock Multi-hop question answering.
\newblock \emph{Foundations and Trends{\textregistered} in Information Retrieval}, 17(5):457--586.

\bibitem[{Min et~al.(2019)Min, Zhong, Zettlemoyer, and Hajishirzi}]{min2019multi}
Sewon Min, Victor Zhong, Luke Zettlemoyer, and Hannaneh Hajishirzi. 2019.
\newblock Multi-hop reading comprehension through question decomposition and rescoring.
\newblock In \emph{ACL}.

\bibitem[{OpenAI(2024)}]{gpt4omini}
OpenAI. 2024.
\newblock \href {https://openai.com/index/gpt-4o-mini-advancing-cost-efficient-intelligence/} {Gpt-4o mini: advancing cost-efficient intelligence}.

\bibitem[{Ouyang et~al.(2022)Ouyang, Wu, Jiang, Almeida, Wainwright, Mishkin, Zhang, Agarwal, Slama, Ray et~al.}]{ouyang2022training}
Long Ouyang, Jeffrey Wu, Xu~Jiang, Diogo Almeida, Carroll Wainwright, Pamela Mishkin, Chong Zhang, Sandhini Agarwal, Katarina Slama, Alex Ray, et~al. 2022.
\newblock Training language models to follow instructions with human feedback.
\newblock \emph{Advances in neural information processing systems}, 35:27730--27744.

\bibitem[{Pezeshkpour and Hruschka(2024)}]{sensitivity1}
Pouya Pezeshkpour and Estevam Hruschka. 2024.
\newblock Large language models sensitivity to the order of options in multiple-choice questions.
\newblock In \emph{Findings of the Association for Computational Linguistics: NAACL 2024}, pages 2006--2017.

\bibitem[{Press et~al.(2023)Press, Zhang, Min, Schmidt, Smith, and Lewis}]{press-etal-2023-measuring}
Ofir Press, Muru Zhang, Sewon Min, Ludwig Schmidt, Noah Smith, and Mike Lewis. 2023.
\newblock \href {https://doi.org/10.18653/v1/2023.findings-emnlp.378} {Measuring and narrowing the compositionality gap in language models}.
\newblock In \emph{Findings of the Association for Computational Linguistics: EMNLP 2023}, pages 5687--5711, Singapore. Association for Computational Linguistics.

\bibitem[{Shi et~al.(2023)Shi, Chen, Misra, Scales, Dohan, Chi, Sch{\"a}rli, and Zhou}]{shi2023large}
Freda Shi, Xinyun Chen, Kanishka Misra, Nathan Scales, David Dohan, Ed~H Chi, Nathanael Sch{\"a}rli, and Denny Zhou. 2023.
\newblock Large language models can be easily distracted by irrelevant context.
\newblock In \emph{International Conference on Machine Learning}, pages 31210--31227. PMLR.

\bibitem[{Shi et~al.(2024{\natexlab{a}})Shi, Min, Yasunaga, Seo, James, Lewis, Zettlemoyer, and Yih}]{shi2024replug}
Weijia Shi, Sewon Min, Michihiro Yasunaga, Minjoon Seo, Richard James, Mike Lewis, Luke Zettlemoyer, and Wen-tau Yih. 2024{\natexlab{a}}.
\newblock Replug: Retrieval-augmented black-box language models.
\newblock In \emph{Proceedings of the 2024 Conference of the North American Chapter of the Association for Computational Linguistics: Human Language Technologies (Volume 1: Long Papers)}, pages 8364--8377.

\bibitem[{Shi et~al.(2024{\natexlab{b}})Shi, Zhang, Sun, Gao, Ren, Chen, and Ren}]{shi-etal-2024-generate-then-ground}
Zhengliang Shi, Shuo Zhang, Weiwei Sun, Shen Gao, Pengjie Ren, Zhumin Chen, and Zhaochun Ren. 2024{\natexlab{b}}.
\newblock \href {https://doi.org/10.18653/v1/2024.acl-long.397} {Generate-then-ground in retrieval-augmented generation for multi-hop question answering}.
\newblock In \emph{Proceedings of the 62nd Annual Meeting of the Association for Computational Linguistics (Volume 1: Long Papers)}, pages 7339--7353, Bangkok, Thailand. Association for Computational Linguistics.

\bibitem[{Springer et~al.(2024)Springer, Kotha, Fried, Neubig, and Raghunathan}]{springer2024repetition}
Jacob~Mitchell Springer, Suhas Kotha, Daniel Fried, Graham Neubig, and Aditi Raghunathan. 2024.
\newblock Repetition improves language model embeddings.
\newblock \emph{arXiv preprint arXiv:2402.15449}.

\bibitem[{Team(2024)}]{qwen2.5}
Qwen Team. 2024.
\newblock \href {https://qwenlm.github.io/blog/qwen2.5/} {Qwen2.5: A party of foundation models}.

\bibitem[{Touvron et~al.(2023)Touvron, Martin, Stone, Albert, Almahairi, Babaei, Bashlykov, Batra, Bhargava, Bhosale et~al.}]{touvron2023llama}
Hugo Touvron, Louis Martin, Kevin Stone, Peter Albert, Amjad Almahairi, Yasmine Babaei, Nikolay Bashlykov, Soumya Batra, Prajjwal Bhargava, Shruti Bhosale, et~al. 2023.
\newblock Llama 2: Open foundation and fine-tuned chat models.
\newblock \emph{arXiv preprint arXiv:2307.09288}.

\bibitem[{Trivedi et~al.(2022)Trivedi, Balasubramanian, Khot, and Sabharwal}]{trivedi2021musique}
Harsh Trivedi, Niranjan Balasubramanian, Tushar Khot, and Ashish Sabharwal. 2022.
\newblock {M}u{S}i{Q}ue: Multihop questions via single-hop question composition.
\newblock \emph{Transactions of the Association for Computational Linguistics}.

\bibitem[{Trivedi et~al.(2023)Trivedi, Balasubramanian, Khot, and Sabharwal}]{trivedi-etal-2023-interleaving}
Harsh Trivedi, Niranjan Balasubramanian, Tushar Khot, and Ashish Sabharwal. 2023.
\newblock \href {https://doi.org/10.18653/v1/2023.acl-long.557} {Interleaving retrieval with chain-of-thought reasoning for knowledge-intensive multi-step questions}.
\newblock In \emph{Proceedings of the 61st Annual Meeting of the Association for Computational Linguistics (Volume 1: Long Papers)}, pages 10014--10037, Toronto, Canada. Association for Computational Linguistics.

\bibitem[{Wei et~al.(2022)Wei, Wang, Schuurmans, Bosma, Xia, Chi, Le, Zhou et~al.}]{wei2022cot}
Jason Wei, Xuezhi Wang, Dale Schuurmans, Maarten Bosma, Fei Xia, Ed~Chi, Quoc~V Le, Denny Zhou, et~al. 2022.
\newblock Chain-of-thought prompting elicits reasoning in large language models.
\newblock \emph{Advances in neural information processing systems}, 35:24824--24837.

\bibitem[{Wu et~al.(2024)Wu, Xie, Chen, Zhu, Zhang, and Xiao}]{wu2024how}
Siye Wu, Jian Xie, Jiangjie Chen, Tinghui Zhu, Kai Zhang, and Yanghua Xiao. 2024.
\newblock \href {https://openreview.net/forum?id=S7NVVfuRv8} {How easily do irrelevant inputs skew the responses of large language models?}
\newblock In \emph{First Conference on Language Modeling}.

\bibitem[{Xiong et~al.(2024)Xiong, Papageorgiou, Lee, and Papailiopoulos}]{xiong2024artificial}
Zheyang Xiong, Vasilis Papageorgiou, Kangwook Lee, and Dimitris Papailiopoulos. 2024.
\newblock From artificial needles to real haystacks: Improving retrieval capabilities in llms by finetuning on synthetic data.
\newblock \emph{arXiv preprint arXiv:2406.19292}.

\bibitem[{Xu et~al.(2024)Xu, Pang, Shen, Cheng, and Chua}]{xu2024search}
Shicheng Xu, Liang Pang, Huawei Shen, Xueqi Cheng, and Tat-Seng Chua. 2024.
\newblock Search-in-the-chain: Interactively enhancing large language models with search for knowledge-intensive tasks.
\newblock In \emph{Proceedings of the ACM on Web Conference 2024}, pages 1362--1373.

\bibitem[{Xu et~al.(2023)Xu, Tao, Shen, Xu, Xu, Long, and Lou}]{xu2023re}
Xiaohan Xu, Chongyang Tao, Tao Shen, Can Xu, Hongbo Xu, Guodong Long, and Jian-guang Lou. 2023.
\newblock Re-reading improves reasoning in language models.
\newblock \emph{arXiv preprint arXiv:2309.06275}.

\bibitem[{Yadav et~al.(2019)Yadav, Bethard, and Surdeanu}]{yadav2019quick}
Vikas Yadav, Steven Bethard, and Mihai Surdeanu. 2019.
\newblock Quick and (not so) dirty: Unsupervised selection of justification sentences for multi-hop question answering.
\newblock In \emph{Proceedings of the 2019 Conference on Empirical Methods in Natural Language Processing and the 9th International Joint Conference on Natural Language Processing (EMNLP-IJCNLP)}, pages 2578--2589.

\bibitem[{Yan et~al.(2024)Yan, Gu, Zhu, and Ling}]{yan2024corrective}
Shi-Qi Yan, Jia-Chen Gu, Yun Zhu, and Zhen-Hua Ling. 2024.
\newblock Corrective retrieval augmented generation.
\newblock \emph{arXiv preprint arXiv:2401.15884}.

\bibitem[{Yang et~al.(2018)Yang, Qi, Zhang, Bengio, Cohen, Salakhutdinov, and Manning}]{yang2018hotpotqa}
Zhilin Yang, Peng Qi, Saizheng Zhang, Yoshua Bengio, William~W. Cohen, Ruslan Salakhutdinov, and Christopher~D. Manning. 2018.
\newblock {HotpotQA}: A dataset for diverse, explainable multi-hop question answering.
\newblock In \emph{Conference on Empirical Methods in Natural Language Processing ({EMNLP})}.

\bibitem[{Yao et~al.(2023)Yao, Zhao, Yu, Du, Shafran, Narasimhan, and Cao}]{yaoreact}
Shunyu Yao, Jeffrey Zhao, Dian Yu, Nan Du, Izhak Shafran, Karthik~R Narasimhan, and Yuan Cao. 2023.
\newblock React: Synergizing reasoning and acting in language models.
\newblock In \emph{The Eleventh International Conference on Learning Representations}.

\bibitem[{Yu et~al.(2023)Yu, Zhang, Pan, Ma, Wang, and Yu}]{yu2023chainofnote}
Wenhao Yu, Hongming Zhang, Xiaoman Pan, Kaixin Ma, Hongwei Wang, and Dong Yu. 2023.
\newblock Chain-of-note: Enhancing robustness in retrieval-augmented language models.
\newblock \emph{arXiv preprint arXiv:2311.09210}.

\bibitem[{Zhao et~al.(2023)Zhao, Zhou, Li, Tang, Wang, Hou, Min, Zhang, Zhang, Dong et~al.}]{zhao2023llmsurvey}
Wayne~Xin Zhao, Kun Zhou, Junyi Li, Tianyi Tang, Xiaolei Wang, Yupeng Hou, Yingqian Min, Beichen Zhang, Junjie Zhang, Zican Dong, et~al. 2023.
\newblock A survey of large language models.
\newblock \emph{arXiv preprint arXiv:2303.18223}.

\bibitem[{Zheng et~al.(2024)Zheng, Zhou, Meng, Zhou, and Huang}]{sensitivity2}
Chujie Zheng, Hao Zhou, Fandong Meng, Jie Zhou, and Minlie Huang. 2024.
\newblock Large language models are not robust multiple choice selectors.
\newblock In \emph{The Twelfth International Conference on Learning Representations}.

\bibitem[{Ziyan~Jiang(2024)}]{jiang2024longrag}
Wenhu~Chen Ziyan~Jiang, Xueguang~Ma. 2024.
\newblock \href {https://arxiv.org/abs/2406.15319} {Longrag: Enhancing retrieval-augmented generation with long-context llms}.
\newblock \emph{arXiv preprint arXiv:2406.15319}.

\end{thebibliography}

\clearpage

\appendix
\section{Proof of Theorem \ref{theorem}} \label{appendix_proofs}
We will prove this theorem by demonstrating that the augmented context $f_{\text{rep}}^{(k)}(C)$, constructed by repeating the original context $C$ $k$ times, inherently arranges the supporting documents in the order specified by the permutation $\sigma$.

By applying the augmentation function, we obtain:

\[
\begin{split}
f_{\text{rep}}^{(k)}(C) = &\ \underbrace{(n_0, d_{\tau(1)}, n_1, \cdots, d_{\tau(k)}, n_k)}_{\text{1st repetition}} \oplus \\
&\ \underbrace{(n_0, d_{\tau(1)}, n_1, \cdots, d_{\tau(k)}, n_k)}_{\text{2nd repetition}} \oplus \\
&\ \cdots \\
&\ \underbrace{(n_0, d_{\tau(1)}, n_1, \cdots, d_{\tau(k)}, n_k)}_{\text{$k$-th repetition}}.
\end{split}
\]

We can extract the supporting documents in the order specified by $\sigma$ by selecting the $d_{\sigma(i)}$ from the $i$-th repetition. All other documents in each repetition of $C$ (i.e., all $d_j$ where $j \neq \sigma(i)$ in the $i$-th repetition and all $n_j$) are considered noisy documents. This means they are treated as $n'_j$ in the definition of $\mathcal{C}_\sigma$. The augmented context $f_{\text{rep}}^{(k)}(C)$ can thus be viewed as:

\[
f_{\text{rep}}^{(k)}(C) = (n'_0, d_{\sigma(1)}, n'_1, d_{\sigma(2)}, \dots, d_{\sigma(k)}, n'_k),
\]

\noindent where each $d_{\sigma(i)}$ is extracted from the $i$-th repetition, and all $n'_j$ are noisy documents. This directly matches the structure of some $C_\sigma \in \mathcal{C}_\sigma$, satisfying:

\[
f_{\text{rep}}^{(k)}(C) \in \mathcal{C}_\sigma.
\]

Since $\sigma$ was arbitrary, this holds for any permutation $\sigma$.

\section{Data Statistics} \label{appendix_data_statics}
Table \ref{tab_data_statistics} shows the statistics of the evaluation sets in the multi-hop QA tasks conducted in the main experiments.

\section{Prompts for Experiments} \label{appendix_prompts}

Table \ref{prompt_mqa_base} and \ref{prompt_synthetic_base} show the prompt template of CoRe for the multi-hop QA tasks and the synthetic task conducted in the section \ref{main_experiments}, respectively.

\section{Additional Results} \label{appendix_addi_results}
\begin{figure*}[]
    \centering
    \includegraphics[width=\textwidth]{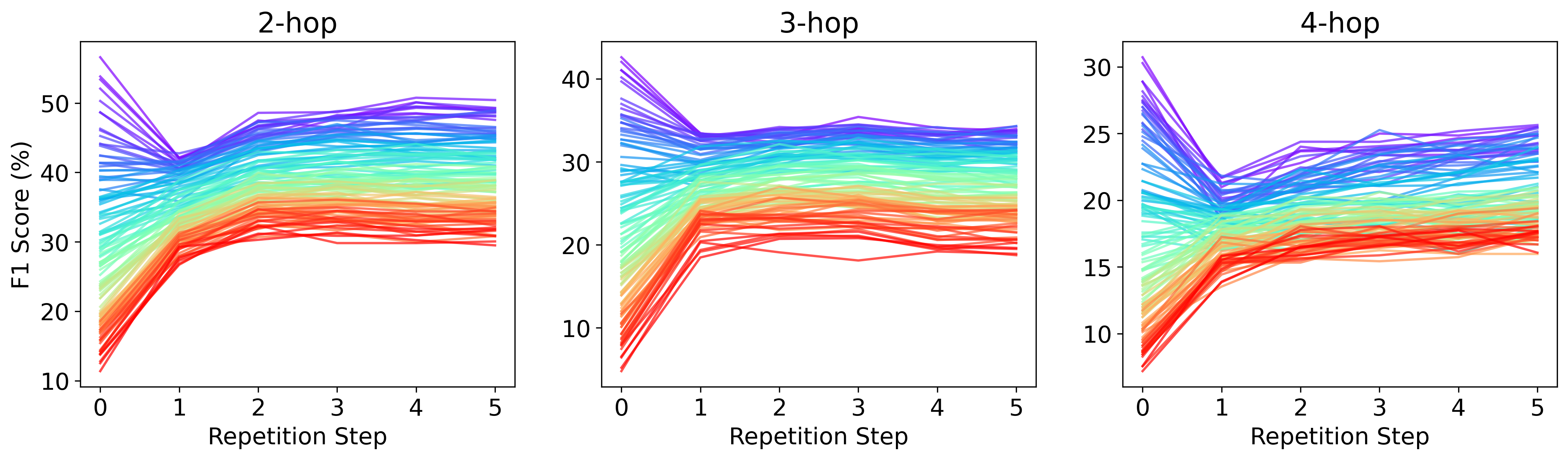} 
    \caption{
    Performance of Mistral-7B-Instruct-v0.3 with permuted contexts of MuSiQue during repetitions. The red line denotes the context in the worst order, and the purple line denotes the context in the best order.
    }
    \label{fig_analysis_permute_mistral}
\end{figure*}

\begin{figure*}[]
    \centering
    \includegraphics[width=\textwidth]{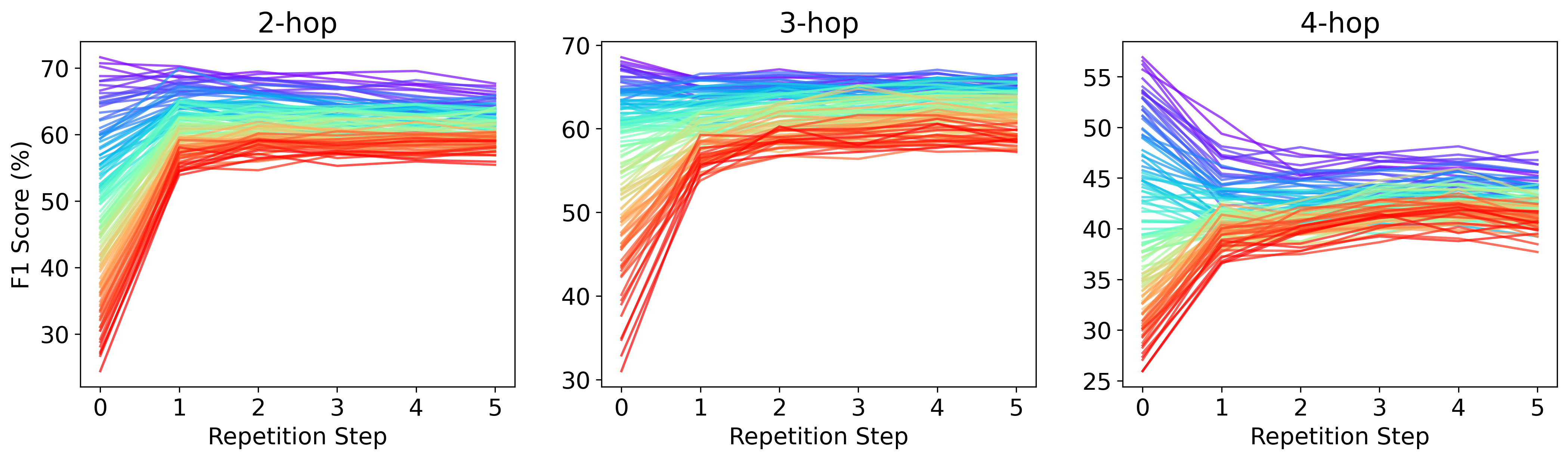} 
    \caption{
    Performance of Qwen2.5-7B-Instruct with permuted contexts of MuSiQue during repetitions. The red line denotes the context in the worst order, and the purple line denotes the context in the best order.
    }
    \label{fig_analysis_permute_qwen}
\end{figure*}

\begin{figure*}[]
    \centering
    \includegraphics[width=\textwidth]{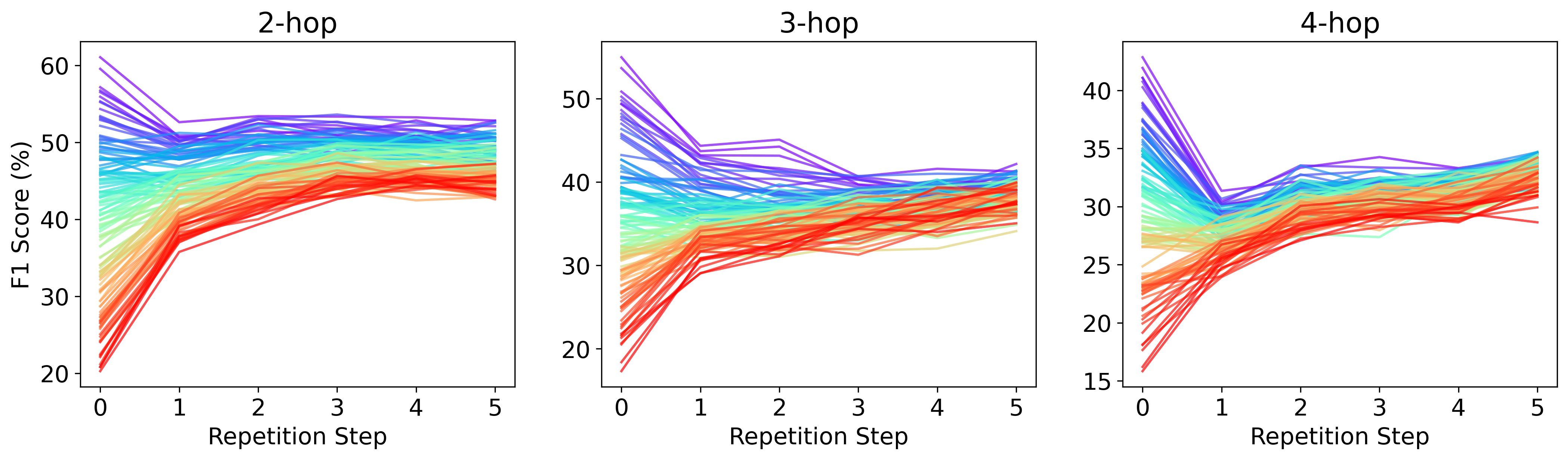} 
    \caption{
    Performance of Phi-3.5-mini-instruct with permuted contexts of MuSiQue during repetitions. The red line denotes the context in the worst order, and the purple line denotes the context in the best order.
    }
    \label{fig_analysis_permute_phi}
\end{figure*}

In this section, we describe additional experimental results not reported in the main text. Specifically, we extend the analysis from Section \ref{sec_4_3}, where the impact of applying CoRe to each permuted context was only evaluated on Llama, to additional models, including Mistral, Qwen, and Phi. Figures \ref{fig_analysis_permute_mistral} to \ref{fig_analysis_permute_phi} present the analysis results for each model. Similar to the findings for Llama, we observe that the performance improvement from CoRe increases as the context be close to the worst-order cases for all models. This further solidifies that the performance gains of CoRe stem from augmenting the original context into an optimal context. Meanwhile, we notice a slight performance drop in contexts close to the best order as repetition increases, particularly for models like Mistral and Phi. This trend is more pronounced compared to Llama and can be interpreted as a result of these models being less capable of handling noisy documents.

Additionally, we evaluate the performance of CoRe when applied to Chain-of-Thought (CoT) reasoning in multi-hop QA tasks. As shown in Table \ref{tab_analysis_cot}, CoRe consistently leads to significant improvements, particularly in bridge and compositional question types, aligning with the main results presented earlier. In the case of GPT-4o-mini, some performance metrics remain stable, which we attribute to an observed trade-off: while recall increases, precision decreases. This suggests that when CoT reasoning is combined with CoRe, the model is better at deducing the correct answer, but the length of the generated response tends to increase, resulting in a slight drop in precision. In conclusion, our results demonstrate that CoRe can meaningfully enhance performance even in the context of CoT reasoning. Table \ref{prompt_mqa_cot} shows the prompt template of CoRe for CoT reasoning in the multi-hop QA tasks.

\section{Experimental Details of Retrieve-and-Reason Task} \label{appendix_detail_retrieve}


In this section, we provide the experimental details of the retrieve-and-reasoning task conducted in Section \ref{sec_5_2}. The retrieval database consists of 101,962 documents, constructed by ensuring no overlap between the documents assigned to each sample from the MuSiQue training and validation sets. The evaluation set is built by randomly sampling 480 samples from the MuSiQue validation set, comprising 249 2-hop samples, 149 3-hop samples, and 82 4-hop samples.

The standard RAG approach retrieves the top 100 documents at once, which are then used as the context for prompting the LLM. IRCoT operates by having the LLM perform Chain-of-Thought (CoT) reasoning while generating sentences. After each sentence is generated, the top 10 documents related to the generated sentence are retrieved and concatenated with the previously retrieved documents, updating the context. If the LLM concludes its reasoning with an expression such as "So the answer is," the subsequent output is taken as the final answer. If the context update process exceeds 10 iterations without reaching a conclusion, the entire output generated during the CoT process is considered the final answer. This setup follows the one described in \citealt{trivedi-etal-2023-interleaving}. When IRCoT is combined with CoRe, after each CoT sentence is generated, the context of the retrieved documents and previously generated CoT responses is repeated in the prompt.

In the question decomposition setting, LLM is instructed to decompose the query into several sub-queries before answering the query. Once the sub-queries are generated, the LLM is directed to perform step-by-step reasoning by following each sub-query, with the process guided by an instruction that ensures the sub-queries are considered during the answering phase. This setting aligns with the prompt design of \citet{press-etal-2023-measuring}. Table \ref{prompt_decompose} shows the prompt template of query decomposition.

\section{Ablation Studies} \label{appendix_ablations}
\subsection{Prompt Style Ablation}
In this section, we investigate the impact of changing the position where CoRe is applied within the prompt. Specifically, we conduct an ablation study for the role where CoRe is applied in the chat template prompt. In our default setup, CoRe is applied through the assistant role, where the context is repeated in the responses generated by the assistant. However, we also consider an alternative setup where the context is repeated in the user role. We evaluate the performance of Llama, Mistral, Qwen, and Phi on 2WikiMultihopQA using F1 scores. $\hat{k}=2$ for CoRe in this study.

The results, presented in Table \ref{tab_ablataion_position}, reveal that in some cases, applying CoRe in the user role leads to significant performance improvements. Notably, except for Qwen, the other models show substantial gains when applying CoRe in the user role. For instance, Mistral demonstrates a 17-point improvement in compositional-type questions and a 15-point gain in comparison-type questions. Phi also shows an 18-point increase in F1 score on bridge-comparison-type questions. We hypothesize that these improvements are due to the task being framed as a single-turn chat when CoRe is applied in the user role. Some models are predominantly tuned on single-turn chat tasks rather than multi-turn interactions, making it easier for them to reason when CoRe is applied in the user role. By adapting the CoRe methodology to align with the model's training environment, we would create a more familiar reasoning context, leading to larger performance gains. Table \ref{prompt_mqa_nochat} shows the prompt template of CoRe applied in the user role position within the prompt.

\subsection{Repetition Style Ablation}
Additionally, we conduct an ablation study to explore the effects of modifying the repeated context. Specifically, we examine 4 cases where the content of the context is altered during repetition: (1) each documents are paraphrased, (2) each documents are summarized, (3) the order of the documents is shuffled randomly, and (4) the order of the documents is reversed. We evaluate the performance of each of these cases. The experiments are conducted on 2WikiMultihopQA using Llama, where the paraphrasing and summarizing are directly performed by Llama responsible for reasoning. $\hat{k}=2$ for CoRe in this study.

Table \ref{tab_ablataion_modify} presents the results for each case when repetition is applied once. Interestingly, in comparison tasks, all ablation settings yield higher performance than the baseline setting, which involves simply repeating the context as is. This suggests that the process of recognizing the same content in different forms is particularly beneficial for comparison tasks. For the cases of shuffling and reversing the order of documents, performance improvements are observed in three out of four types, excluding inference. This indicates that varying the order of documents during repetition enables the model to consider a wider range of reasoning chains. As discussed in Section \ref{sec_4_3}, future research could explore applying CoRe while reducing the noise power within the context, and actively modifying the form of the context, as in this study, could be another potential direction to consider. Table \ref{prompt_paraphrase} and \ref{prompt_summary} show the prompt template of paraphrasing and summarizing, respectively.

\section{Computational Environment} \label{appendix_compute_environments}
For the main experiments, we utilize NVIDIA A40 GPU. For the experiments for the retrieve-and-reason task, we utilize NVIDIA H100 80GB HBM3 GPU. All the models except GPT-4o-mini in our work are from \texttt{huggingface}. We utilize GPT-4o-mini through \texttt{OpenAI} API package.

\section{Licenses} \label{appendix_license}
HotpotQA, 2WikiMultihopQA, and MuSiQue are licensed under the CC BY-SA 4.0, CC-BY-4.0, and Apache-2.0 licenses, respectively. Llama-3.1, Mistral, Qwen, Phi, and GPT-4o-mini are licensed under the Llama 3.1 COMMUNITY LICENSE, Apache-2.0,  Apache-2.0, MIT, and OpenAI.

\section{Usage of AI Writing Assistant} \label{appendix_ai_writing}
This paper benefited from linguistic support provided by the AI assistant ChatGPT-4o, which contributed by offering paraphrasing and enhancing the original content. No other assistance was sought beyond these services.
\begin{table*}[!t]
    \centering
    \resizebox{\textwidth}{!}{
        \renewcommand{\arraystretch}{1.1}
        \begin{tabular}{l||cc|cccc|ccc}
            \toprule
            \multicolumn{1}{c||}{\multirow{2}{*}{\textbf{Datasets}}} & \multicolumn{2}{c|}{HotpotQA} & \multicolumn{4}{c|}{2WikiMultihopQA} & \multicolumn{3}{c}{MuSiQue}                                                                                                          \\
            \multicolumn{1}{c||}{}                                 & Comparison                             & Bridge                                        & Compositional                        & Comparison     & Inference      & Bridge-Comparison & 2-hop          & 3-hop          & 4-hop          \\ \hline
            \textbf{\# of samples}                                         & 1000& 1000& 500& 500& 500& 500& 1252& 760& 405\\
            \textbf{Context length}                                         & 869.17& 977.33& 701.15& 567.84& 772.21& 593.87& 1694.4& 1827.92& 1778.8\\
            \bottomrule
        \end{tabular}
    }
    \caption{Data statistics of evaluation sets for each dataset of multi-hop QA tasks in the main experiments. Context length denotes the mean number of words included in the context of each type of sample.}
    \label{tab_data_statistics}
\end{table*}

\begin{table*}
\centering
{\small
\renewcommand{\arraystretch}{1.2}
\begin{tabular}{p{15cm}}
\toprule
\texttt{\textbf{[User role]}} \\
\texttt{Question:} \{\textit{Question}\} \\ \\
\texttt{Documents:}\\ 
\texttt{Document [0]} \{\textit{Document 0}\} \\ 
\texttt{Document [1]} \{\textit{Document 1}\} \\ 
... \\
\texttt{Document [N]} \{\textit{Document N}\} \\ \\
\texttt{Answer the question based on the given documents. Respond only the answer within a few words after 'Answer:'.} \\ \\

\texttt{\textbf{[Assistant role]}} \\
\texttt{Sure. Before answering the question, I'll reconsider the question and the documents once more.} \\ \\
\texttt{Question:} \{\textit{Question}\} \\ \\
\texttt{Documents:}\\ 
\texttt{Document [0]} \{\textit{Document 0}\} \\ 
\texttt{Document [1]} \{\textit{Document 1}\} \\ 
... \\
\texttt{Document [N]} \{\textit{Document N}\} \\ \\

\texttt{\textbf{[User role]}} \\
\texttt{Now answer the question based on the documents. Respond only the answer within a few words after 'Answer:'.} \\ \\

\texttt{\textbf{[Assistant role]}} \\
\texttt{Answer:}
\\ \bottomrule
\end{tabular}
}
\caption{Prompt template of CoRe used in main experiments for the multi-hop QA tasks.} \label{prompt_mqa_base}
\end{table*}
\begin{table*}
\centering
{\small
\renewcommand{\arraystretch}{1.2}
\begin{tabular}{p{15cm}}
\toprule
\texttt{\textbf{[User role]}} \\
\texttt{Information:} \\
\texttt{All the \{\textit{k}\} lists described in the below contain exactly \{\textit{n}\} elements.} \\
\texttt{In the list 0,  \{\textit{lists[0][0]}\} is positioned immediately before  \{\textit{lists[0][1]}\}.} \\
... \\
\texttt{In the list \{\textit{k}\},  \{\textit{lists[k][0]}\} is positioned immediately before  \{\textit{lists[k][1]}\}.} \\
... \\
\texttt{In the list 0, \{\textit{lists[0][n-2]}\} is positioned immediately before \{\textit{lists[0][n-1]}\}.} \\
... \\
\texttt{In the list \{\textit{k}\},  \{\textit{lists[k][n-2]}\} is positioned immediately before  \{\textit{lists[k][n-1]}\}.} \\ \\
\texttt{Question: What is the first element of the list that contains \{\textit{lists[0][n-1]}\}?}

\texttt{Answer the question based on the given information. Respond only the answer without any explanation after 'Answer:'.} \\ \\

\texttt{\textbf{[Assistant role]}} \\
\texttt{Sure. Before answering the question, I'll reconsider the question and the documents \{\textit{t}\} times more.} \\ \\
\texttt{Information:} \\
\texttt{\{\textit{Information}\}} \\ \\
\texttt{Question:} \texttt{\{\textit{Question}\}} \\
... \\ (repeat t-times) \\ \\
\texttt{\textbf{[User role]}} \\
\texttt{Now answer the question based on the documents. Respond only the answer without any explanation after 'Answer:'.} \\ \\

\texttt{\textbf{[Assistant role]}} \\
\texttt{Answer:}
\\ \bottomrule
\end{tabular}
}
\caption{Prompt template of CoRe used in main experiments for the synthetic task.} \label{prompt_synthetic_base}
\end{table*}

\begin{table*}[!t]
    \centering
    \resizebox{\textwidth}{!}{
        \renewcommand{\arraystretch}{1.1}
        \begin{tabular}{l||cc|cccc|ccc}
            \toprule
            \multicolumn{1}{c||}{\multirow{2}{*}{\textbf{Models}}} & \multicolumn{2}{c|}{\textbf{HotpotQA}} & \multicolumn{4}{c|}{\textbf{2WikiMultihopQA}} & \multicolumn{3}{c}{\textbf{MusiQue}}                                                                                                          \\
            \multicolumn{1}{c||}{}                                 & Comparison                             & Bridge                                        & Compositional                        & Comparison     & Inference      & Bridge-Comparison & 2-hop          & 3-hop          & 4-hop          \\ \hline
            Llama-3.1-8B                                           & \textbf{54.07}                         & 67.74                                         & 52.36                                & 70.46          & 61.21          & 42.23             & 49.71          & 43.03          & 31.28          \\
            + CoRe ($\hat{k}=2$)                                                & 53.07                                  & \textbf{71.93}                                & \textbf{53.36}                       & \textbf{72.12} & \textbf{63.54} & \textbf{77.55}    & \textbf{51.94} & \textbf{44.7}  & \textbf{37.98} \\ \hline
            Mistral-7B                                             & \textbf{32.19}                         & 43.68                                         & 27.31                                & 39.74          & 37.22          & 26.47             & 21.87          & 14.18          & 9.41           \\
            + CoRe ($\hat{k}=2$)                                                  & 36.04                                  & \textbf{52.74}                                & \textbf{37.37}                       & \textbf{45.94} & \textbf{48.51} & \textbf{53.47}    & \textbf{25.99} & \textbf{15.96} & \textbf{11.22} \\ \hline
            Qwen2.5-7B                                             & 65.93                                  & 68.01                                         & 49.29                                & 84.83          & 55.37          & 77.38             & 45.33          & 38.18          & 36.53          \\
            + CoRe ($\hat{k}=2$)                                                  & \textbf{65.67}                         & \textbf{72.03}                                & \textbf{51.09}                       & \textbf{85.59} & \textbf{59.97} & \textbf{83.96}    & \textbf{49.94} & \textbf{45.21} & \textbf{40.25} \\ \hline
            Phi-3.5-mini                                           & \textbf{44.58}                         & 59.76                                         & 44.92                                & 81.44          & 48.48          & 81.14             & 28.53          & 23.39          & 19.49          \\
            + CoRe ($\hat{k}=2$)                                                  & 46.23                                  & \textbf{62.4}                                 & \textbf{46.01}                       & \textbf{83.26} & \textbf{51.23} & \textbf{82.83}    & \textbf{34.98} & \textbf{31.24} & \textbf{30.61} \\ \hline
            GPT-4o-mini                                            & \textbf{59.29}                         & 71.74                                         & \textbf{55.75}                       & \textbf{78.91} & 76.44          & \textbf{88.46}    & 55.06          & 49.68          & 41.84          \\
            + CoRe ($\hat{k}=2$)                                                 & 57.68                                  & \textbf{74.04}                                & 54.93                                & 77.53          & \textbf{77.65} & 88.38             & \textbf{56.8}  & \textbf{51.41} & \textbf{44.35} \\
            \bottomrule
        \end{tabular}
    }
    \caption{Chain-of-Thought results of F1 score in the multi-hop QA tasks. All models are instruction-tuned LLMs.}
    \label{tab_analysis_cot}
\end{table*}

\begin{table*}
\centering
{\small
\renewcommand{\arraystretch}{1.2}
\begin{tabular}{p{15cm}}
\toprule
\textbf{Chain-of-Thought reasoning} \\ \\
\texttt{\textbf{[User role]}} \\
\texttt{Question:} \{\textit{Question}\} \\ \\
\texttt{Documents:}\\ 
\texttt{Document [0]} \{\textit{Document 0}\} \\ 
\texttt{Document [1]} \{\textit{Document 1}\} \\ 
... \\
\texttt{Document [N]} \{\textit{Document N}\} \\ \\
\texttt{Answer the question based on the given documents.} \\ \\

\texttt{\textbf{[Assistant role]}} \\
\texttt{Sure. Before answering the question, I'll reconsider the question and the documents once more.} \\ \\
\texttt{Question:} \{\textit{Question}\} \\ \\
\texttt{Documents:}\\ 
\texttt{Document [0]} \{\textit{Document 0}\} \\ 
\texttt{Document [1]} \{\textit{Document 1}\} \\ 
... \\
\texttt{Document [N]} \{\textit{Document N}\} \\ \\

\texttt{\textbf{[User role]}} \\
\texttt{Now answer the question based on the documents.} \\ \\

\texttt{\textbf{[Assistant role]}} \\
\texttt{Let's think step by step.}
\\ \hline
\textbf{Extract short answer from CoT response} \\ \\

(After same prompt of user-assistant-user as above prompt for chain-of-thought reasoning ...) \\ \\

\texttt{\textbf{[Assistant role]}} \\
\texttt{Let's think step by step.} \{\textit{CoT response}\}  \\ \\ 

\texttt{\textbf{[User role]}} \\
\texttt{Respond only the answer in a few words after 'Answer:'.} \\ \\

\texttt{\textbf{[Assistant role]}} \\
\texttt{Answer:}
\\ \bottomrule
\end{tabular}
}
\caption{Prompt template of CoRe used in Chain-of-Thought reasoning for the multi-hop QA tasks.} \label{prompt_mqa_cot}
\end{table*}

\begin{table*}
\centering
{\small
\renewcommand{\arraystretch}{1.2}
\begin{tabular}{p{15cm}}
\toprule
\texttt{\textbf{[User role]}} \\
\texttt{Decompose the following question into several sub-questions.} \\ 
\texttt{Question:} \texttt{\{\textit{Question}\}} \\ \\ 

\texttt{\textbf{[Assistant role]}} \\
\texttt{1. }
\\ \bottomrule
\end{tabular}
}
\caption{Prompt template of query decomposition for the retrieve-and-reason task.} \label{prompt_decompose}
\end{table*}

\begin{table*}[!t]
    \centering
    \resizebox{0.8\textwidth}{!}{
        \begin{tabular}{l|cccc}
            \toprule
            \multicolumn{1}{c|}{\textbf{Models}} & \textbf{Compositional} & \textbf{Comparison} & \textbf{Inference} & \textbf{Bridge-Comparison} \\ \hline
            Llama-3.1-8B                         & 38.1                   & 55.77               & 40.75              & 34.86                      \\
            + CoRe in assistant                  & 48.15                  & 58.13               & 44.81              & 65.11                      \\
            + CoRe in user                       & \textbf{53.45}         & \textbf{69.36}      & \textbf{47.59}     & \textbf{68.76}             \\ \hline
            Mistral-7B                           & 18.82                  & 32.9                & 29.77              & 15.43                      \\
            + CoRe in assistant                  & 27.78                  & 35.67               & 36.39              & \textbf{29.09}             \\
            + CoRe in user                       & \textbf{44.07}         & \textbf{50.7}       & \textbf{39.64}     & 25.29                      \\ \hline
            Qwen2.5-7B                           & 41.36                  & 65.47               & 37.26              & 39.45                      \\
            + CoRe in assistant                  & \textbf{54.2}          & \textbf{70.3}       & \textbf{42.4}      & 64.54                      \\
            + CoRe in user                       & 54.05                  & 69.73               & 41.64              & \textbf{68.3}              \\ \hline
            Phi-3.5-mini                         & 27.7                   & 45.26               & 27                 & 35.19                      \\
            + CoRe in assistant                  & 37.64                  & 50.92               & 32.34              & 36.71                      \\
            + CoRe in user                       & \textbf{42.56}         & \textbf{54.89}      & \textbf{34.51}     & \textbf{54.37}             \\
            \bottomrule
        \end{tabular}
    }
    \caption{Results in the ablation study about the position of context repetition}
    \label{tab_ablataion_position}
\end{table*}

\begin{table*}[!t]
    \centering
    \resizebox{0.8\textwidth}{!}{
        \begin{tabular}{l|cccc}
            \toprule
            \multicolumn{1}{c|}{\textbf{Methods}} & \textbf{Compositional} & \textbf{Comparison} & \textbf{Inference} & \textbf{Bridge-Comparison} \\ \hline
            Baseline (CoRe)                              & 48.15                  & 58.13               & \textbf{44.81}     & 65.11                      \\
            Paraphrase                            & 48.23                  & 61.17               & 42.32              & 57.19                      \\
            Summary                               & 48.44                  & 61.68               & 40.35              & 64.79                      \\
            Shuffle                               & \textbf{51.25}         & \textbf{63.4}       & 43.33              & 65.97                      \\
            Reverse                               & 50.65                  & 63.35               & 42.9               & \textbf{67.11}             \\
            \bottomrule
        \end{tabular}
    }
    \caption{Results for Llama in the ablation study about the content of repeated contexts.}
    \label{tab_ablataion_modify}
\end{table*}

\begin{table*}
\centering
{\small
\renewcommand{\arraystretch}{1.2}
\begin{tabular}{p{15cm}}
\toprule
\texttt{\textbf{[User role]}} \\
\texttt{Question:} \{\textit{Question}\} \\ \\
\texttt{Documents:}\\ 
\texttt{Document [0]} \{\textit{Document 0}\} \\ 
\texttt{Document [1]} \{\textit{Document 1}\} \\ 
... \\
\texttt{Document [N]} \{\textit{Document N}\} \\ \\
\texttt{Answer the question based on the given documents. Respond only the answer within a few words after 'Answer:'.} \\ \\

\texttt{Look again the input prompt:} \\ \\ 
\texttt{Question:} \{\textit{Question}\} \\ \\
\texttt{Documents:}\\ 
\texttt{Document [0]} \{\textit{Document 0}\} \\ 
\texttt{Document [1]} \{\textit{Document 1}\} \\ 
... \\
\texttt{Document [N]} \{\textit{Document N}\} \\ \\
\texttt{Answer the question based on the given documents. Respond only the answer within a few words after 'Answer:'.} \\ \\
\texttt{\textbf{[Assistant role]}} \\
\texttt{Answer:}
\\ \bottomrule
\end{tabular}
}
\caption{Prompt template of CoRe applied in the user role position.} \label{prompt_mqa_nochat}
\end{table*}
\begin{table*}
\centering
{\small
\renewcommand{\arraystretch}{1.2}
\begin{tabular}{p{15cm}}
\toprule
\texttt{\textbf{[System role]}} \\
\texttt{You are a professional paraphraser. Your task is to paraphrase the given text based on the below instructions. Follow the instructions to achieve the desired output.} \\ \\ 

\texttt{- Objective: Rewrite the text more thoroughly, changing both vocabulary and sentence structure while preserving the original meaning.} \\
\texttt{- Instructions:}
\texttt{(MOST IMPORTANT) Use your own style for natural paraphrase} \\
\texttt{Introduce new expressions, alter sentence structure, and rearrange clauses.} \\
\texttt{Use synonyms and change the form of words (e.g., verbs to nouns, or active to passive voice).} \\
\texttt{Retain the original message but express it in a noticeably different way.} \\ \\

\texttt{- Example:} \\
\texttt{Original: "The quick brown fox jumps over the lazy dog."} \\
\texttt{Paraphrase: "With swift movements, the brown fox leaps over the dog lying lazily."} \\ \\

\texttt{\# Format of the paraphrasing task} \\
\texttt{- Original: The original text.} \\
\texttt{- Paraphrase: The paraphrased version of the text based on the above guideline. Provide the output text immediately.} \\ \\

\texttt{\textbf{[User role]}} \\
\texttt{Paraphrase the original text below.} \\ 
\texttt{Original:} \texttt{\{\textit{input text}\}} \\ \\ 

\texttt{\textbf{[Assistant role]}} \\
\texttt{Paraphrase:}
\\ \bottomrule
\end{tabular}
}
\caption{Prompt template of paraphrasing for the ablation study.} \label{prompt_paraphrase}
\end{table*}
\begin{table*}
\centering
{\small
\renewcommand{\arraystretch}{1.2}
\begin{tabular}{p{15cm}}
\toprule
\texttt{\textbf{[System role]}} \\
\texttt{ummarize the following text while ensuring that no key information, factual accuracy, or essential meaning is lost. Follow these guidelines:} \\ \\ 

\texttt{- Maintain Key Details: All critical points, facts, and arguments from the original text must be preserved.} \\
\texttt{- Conciseness: The summary should be significantly shorter than the original text while capturing its essence.}
\texttt{- Clarity and Precision: Use clear, professional language. Avoid vague phrasing.} \\
\texttt{- No Alteration of Meaning: Do not add, alter, or infer information that is not present in the original text.} \\ \\

\texttt{Please follow below pattern of example. Provide the output text immediately.
} \\ \\

\texttt{- Example:} \\
\texttt{Original Text: The rapid development of artificial intelligence over the last decade has led to significant breakthroughs in various fields, including healthcare, finance, and transportation. However, these advancements also raise concerns about data privacy, job displacement, and the ethical use of AI technologies.} \\
\texttt{Summary: AI advancements in healthcare, finance, and transportation have been substantial, though concerns about data privacy, job displacement, and ethical issues have emerged.} \\ \\

\texttt{\textbf{[User role]}} \\
\texttt{Summarize the following text below.} \\ 
\texttt{Original Text:} \texttt{\{\textit{input text}\}} \\ \\ 

\texttt{\textbf{[Assistant role]}} \\
\texttt{Summary:}
\\ \bottomrule
\end{tabular}
}
\caption{Prompt template of summarizing for the ablation study.} \label{prompt_summary}
\end{table*}

\end{document}